\documentclass[10pt]{article}

\PassOptionsToPackage{colorlinks=true,citecolor=blue,linkcolor=blue,urlcolor=blue}{hyperref}
\usepackage[preprint]{tmlr}

\usepackage{amsmath}
\usepackage{amssymb}
\usepackage{amsthm}
\usepackage{bm}
\usepackage{booktabs}
\usepackage{graphicx}
\usepackage{xcolor}
\usepackage{hyperref}
\usepackage{url}

\newtheorem{theorem}{Theorem}
\newtheorem{proposition}{Proposition}
\newtheorem{lemma}{Lemma}
\newtheorem{corollary}{Corollary}
\theoremstyle{definition}
\newtheorem{definition}{Definition}
\newtheorem{assumption}{Assumption}
\newtheorem{example}{Example}
\theoremstyle{remark}
\newtheorem{remark}{Remark}

\newcommand{\ghat}{\hat{g}}                 %
\newcommand{\etath}{\eta}                   %
\newcommand{\etalam}{\alpha}                %
\newcommand{\epsth}{\varepsilon_\theta}     %
\newcommand{\epslam}{\varepsilon_\lambda}   %
\newcommand{\epscrit}{\varepsilon_{\mathrm{crit}}}
\newcommand{\wnat}{\tilde{\omega}_0}        %
\newcommand{\wex}{\omega_0}                 %
\newcommand{\rratio}{r}                     %
\newcommand{\proj}[1]{\left[#1\right]_{+}}

\newcommand{\TABLE}[3]{\caption{#1}\centering #2\par\smallskip{\footnotesize #3\par}}
\newcommand{\FIGURE}[3]{\centering #1\caption{#2}\par\smallskip{\footnotesize #3\par}}
\newcommand{\up}{\rule{0pt}{2.4ex}}         %
\newcommand{\down}{\rule[-1.2ex]{0pt}{0pt}} %
\renewenvironment{proof}[1]{\par\smallskip\noindent\textit{#1.}\enspace}{\par\smallskip}
\makeatletter
\def\Halmos{\leavevmode\unskip\penalty9999 \hbox{}\nobreak\hfill\quad\hbox{$\square$}}
\makeatother

\def\month{MM}
\def\year{YYYY}
\def\openreview{\url{https://openreview.net/forum?id=XXXX}}

\title{Parametric Resonance in Primal-Dual Learning\\for Slowly Non-Stationary Constrained MDPs}

\author{\name Maxime Baelde \email maxime@antvoice.com \\
      \addr Antvoice, Paris, France}

\begin{document}

\maketitle
\begin{abstract}
Primal-dual methods for constrained Markov decision processes are analyzed under a two-timescale assumption, with non-stationarity entering through two channels. In the \emph{additive} channel, where the constraint level oscillates, the dual oscillation saturates at a floor set by the moving optimum and is amplified linearly by the timescale ratio. In the \emph{parametric} channel, where the constraint sensitivity oscillates, a \emph{coherent} modulation carries no resonant forcing at any depth: the dual update's stiffness modulation cancels the primal update's damping modulation. Instability instead comes from a \emph{mismatch} between the constraint estimate used by the primal update and the measurement used by the dual update, producing a parametric resonance at twice the pair's natural frequency. The tolerated mismatch grows with the square root of the forcing bandwidth and collapses with the square root of the dual-to-primal step-size ratio. These results hold on the tangent linear-quadratic reduction around an active saddle, for one constraint and for $K$ constraints under a common modulation and an isotropy condition. Driving both updates from the same estimate removes the resonant forcing but introduces a stationary offset equal to the estimator's bias. In a Lagrangian soft actor-critic agent, the multiplier locks onto the sub-harmonic at the timescale ratios its identified linear loop predicts and a coherent cost critic suppresses the lock-in, while a proportional (PID) term, predicted to leave the lock-in in place, suppresses most of it through the projection of the multiplier.
\end{abstract}

\section{Introduction}\label{sec:intro}

Constrained Markov decision processes \citep{altman1999} are the standard model for sequential decision problems in which a return has to be maximized subject to a long-run budget, risk or safety requirement. The dominant solution method, alongside the trust-region approach of \citet{achiam2017}, is Lagrangian relaxation: a multiplier weights the constraint, the policy is improved against the resulting unconstrained objective, and the multiplier is driven by the observed constraint violation. The resulting primal-dual loop is what almost every deployed constrained learner actually runs, from safe reinforcement learning benchmarks \citep{ray2019} and libraries \citep{omnisafe2024} to budget pacing in industrial recommendation and real-time bidding \citep{balseiro2019,balseiro2020}. Its appeal is that each half is a familiar object, a policy gradient step and a projected subgradient step, with convergence argued by placing the two on separate timescales \citep{borkar2008,borkar2005,bhatnagar2012}.

The application that motivates this paper is budget pacing in real-time bidding. An agent bids on impressions and must spend a target budget over a day, so the constraint is a spend rate and the multiplier is the shadow price of budget. The environment is non-stationary on a daily cycle: traffic volume, competitive pressure and conversion propensity all move by large factors between night and evening peak. Convergence in a stationary environment settles nothing here; the practitioner's question is whether the multiplier tracks the moving constrained optimum or oscillates around it, and what the learning rates have to do with the answer.

Two expectations recur in practice. The first is that a slowly varying environment can be handled by making the dual update slow enough to average over it. The second is that the timescale separation invoked to justify the analysis also makes the loop robust. Both fail on the tangent linear-quadratic reduction that the rest of the paper analyzes, defined in Section~\ref{sec:setup_cmdp}. The non-stationarity of the constraint \emph{level} produces a bounded oscillation with a floor that no step size removes, and the non-stationarity of the constraint \emph{sensitivity} carries no resonant forcing at all as long as the primal and the dual see the same sensitivity. What destabilizes the loop is the mismatch between two estimates of the same quantity, an implementation property internal to the algorithm, and the timescale separation is what amplifies it.

Constrained reinforcement learning writes a single symbol for both channels: the primal update consumes a constraint \emph{estimate}, differentiable in the policy parameter, and the dual update consumes a constraint \emph{measurement}. Section~\ref{sec:related} positions this split against the feedback-optimization and CMDP literatures; what neither considers is a periodic mismatch between the two objects, which is what this paper adds.

We separate the two channels formally in Definition~\ref{def:channels}. In the additive channel the constraint level oscillates and the linearized dynamics stay time-invariant. In the parametric channel the constraint sensitivity oscillates, the coefficients become time-periodic, and the loop admits exponential growth once the mismatch crosses a threshold set by the damping; this is the realistic channel in real-time bidding (Section~\ref{sec:setup_channels}). Aligning the two objects is the design prescription that follows: if a single constraint estimate drives both updates, the resonant forcing vanishes by construction at every modulation depth. Alignment is impossible when the dual has to be driven by an audited measurement while the primal needs a differentiable surrogate. In that case the tolerated mismatch scales as the square root of the ratio of the primal to the dual step size. Lowering the dual step size therefore widens the tolerance, but it also slows the multiplier's tracking of a moving constraint.

\label{sec:contributions}
Notation is collected in Table~\ref{tab:notation} (Appendix~\ref{sec:notation}). The central fact is structural and holds to first order on the continuous surrogate of the coupled dynamics: on the tangent linear-quadratic reduction of a $K=1$ CMDP around an active constrained saddle point, the primal update modulates damping and stiffness together while the dual update modulates stiffness alone (Lemma~\ref{lem:cancellation}). A common modulation of both channels therefore cancels exactly at the principal tongue, and only the mismatch $\epslam - \epsth$ survives as an effective forcing depth. Three results follow on this reduction, and the resonance follows from classical tools applied to that effective depth; Section~\ref{sec:experiments} compares each with the affine recursion.

\begin{itemize}
\item[(R1)] \textbf{Exact tracking of a moving optimum} (Proposition~\ref{prop:floor}(i)). The DC gain from constraint disturbance to multiplier is exactly $a/(\etalam c^2)$, precisely what is needed to track the instantaneous optimal multiplier $\lambda^\star(t) = ab_t/c^2$. The peak-to-peak excursion $2Aa/c^2$ is therefore the size of that moving target, independently of both step sizes. This refutes, in the coupled system, the natural extrapolation of the frozen-policy bound of \citet{baelde2026companion}, which grows as $\mathcal O(\etalam/\omega)$ as the forcing slows down.
\item[(R2)] \textbf{Worst-case amplification by the timescale separation} (Proposition~\ref{prop:worstcase}(ii)). Maximizing over the forcing frequency, the peak-to-peak dual oscillation is $A\etalam/\sigma$, reducing at leading order to $(2Aa/c^2)\max\{1, (2\zeta)^{-2}\}$ with $\zeta$ the reduced damping and $(2\zeta)^{-2} = \rratio c^2/a^2$. Past $\rratio = a^2/c^2$, growing timescale separation amplifies the worst case linearly in $\rratio$.
\item[(R3)] \textbf{Coherent sensitivity modulation cannot destabilize the loop} (Proposition~\ref{thm:coherent}). If the primal and the dual see the same modulated sensitivity the resonant forcing vanishes identically at every depth, waveform and frequency. The weighted energy $\etalam\theta^2 + \etath\lambda^2$ is then a Lyapunov function for the continuous flow, and the top Floquet exponent is strictly negative without any smallness assumption. What remains is a damping-borne pump bounded by $\sigma J_1(\varepsilon)$, giving exponent $-\sigma(1-J_1(\varepsilon))$ at the principal tongue. The recursion stays stable at every admissible depth under a condition on the step sizes alone.
\end{itemize}

Section~\ref{sec:parametric} sharpens (R2) into a quantitative threshold: Theorem~\ref{thm:threshold} gives the narrowband mismatch threshold $|\epslam-\epsth| > 4\zeta$ at which the resonance turns unstable, Corollary~\ref{cor:existence} restates it as a condition on the step sizes alone, $\rratio > (a/c)^2$, and Proposition~\ref{prop:spectral} extends it to a mismatch spread over a bandwidth $B$, with a square-root law in $B$. Section~\ref{sec:exp_deeprl} carries the predictions to a Lagrangian soft actor-critic agent: the multiplier locks onto the sub-harmonic at the timescale ratios the agent's identified linear loop predicts, a coherent cost critic suppresses the lock-in, and, contrary to its linear prediction, the proportional term of a PID Lagrangian agent suppresses most lock-ins through the saturation of the multiplier's projection while leaving the forced response and most of the return loss in place.

These statements hold on the tangent reduction, not for general CMDPs: Section~\ref{sec:setup_cmdp} states what the scalar reduction assumes, the projection in \eqref{eq:dual} is inactive throughout the analysis and enters only the experiments, and Section~\ref{sec:scope} delimits the scope further.

\section{Related work}\label{sec:related}

The continuous-time primal-dual method itself is due to \citet{arrow1958}; \citet{platt1988} write it as a damped second-order oscillator, a reduction that \citet{stooke2020} build explicitly on to motivate PID Lagrangian control: eliminating the multiplier from $\dot x = -\nabla f - \lambda\nabla g$ and $\dot\lambda = \etalam g(x)$ yields $\ddot x + D\dot x + \etalam g(x)\nabla g = 0$ with $D = \nabla^2 f + \lambda\nabla^2 g$, an oscillator whose damping is a weighted sum of Hessians. \citet{feijer2010} analyze the stability of exactly this class of primal-dual gradient dynamics via Lyapunov and invariance arguments, characterizing when the undamped oscillatory regime persists. What that lineage writes is a \emph{single} constraint function consumed by both updates.

A second lineage, feedback and time-varying optimization, already keeps the two objects separate. \citet{bernstein2019} and \citet{colombino2020} drive a primal-dual recursion from a model-based gradient while a measurement of the controlled plant enters through feedback, surveyed as a general design principle by \citet{hauswirth2024}. \citet{simonetto2020} and \citet{tang2022} bound the tracking error of running primal-dual and gradient methods against a time-varying optimum under the same estimate/measurement asymmetry. Reward-constrained policy optimization \citep{tessler2019} shares that same split from the constrained-RL side, shaping the reward against a differentiable surrogate cost while updating the multiplier slowly from a Monte Carlo return, an estimate/measurement pair rather than a continuous feedback loop. None of these papers considers a \emph{periodic} mismatch between the two channels, or exhibits the resonance it opens in a primal-dual recursion with inertia. Where a tracking guarantee is proved, as in \citet{simonetto2020} and \citet{tang2022}, it bounds the error against a drifting optimum and leaves the sign of a Floquet exponent unaddressed. \citet{hauswirth2024} surveys the design principle without proving such a bound. What we add is dynamical: periodicity turns the mismatch structure just described into a resonant forcing on the oscillator introduced above. The proportional term of PID Lagrangian control acts on that same oscillator by adding damping, mitigating a resonant forcing once it exists without removing its source.

The convergence mechanism of the canonical two-timescale theorem \citep{borkar2008} is a quasi-static argument, comparing the fast variable's interpolated trajectory to that of the frozen-coefficient ODE (the same ODE-comparison mechanism underlies the two-timescale stochastic approximation of \citet{benveniste1990}). Stability of a frozen-coefficient family does not imply stability of the time-varying system: the textbook counterexample is \citet{markus1960}'s matrix $A(t)$, with spectrum in the open left half-plane at every fixed $t$ yet an unstable $\dot x = A(t)x$ \citep{khalil2002}. \citet{rohrs1985} document the same failure mode in adaptive control: a loop that is stable under every frozen-parameter analysis can burst into instability once unmodeled high-frequency dynamics are excited by persistent, structured input. Parametric resonance in the primal-dual loop is an instance of that known phenomenon. What remains to supply is which physical quantity opens the tongue here, why the coherent case is protected by an exact cancellation, and what the criterion becomes under broadband forcing. Extremum-seeking control deliberately engineers a related resonance, superposing a periodic dither on a slow adaptation loop and analyzing its stability \citep{krstic2000}; there the periodicity is the control input, here an exogenous property of the environment. The line of work descending from \citet{borkar2005}, continued by \citet{bhatnagar2012} and \citet{panda2023}, places the multiplier on a timescale slower than the primal, the side on which the pair has no oscillatory mode to resonate through. Section~\ref{sec:setup_timescales} shows that deployed implementations sit on the other side. \citet{paternain2019} establish zero duality gap for the constrained policy optimization problem itself, a guarantee on the value attained at the optimum. It says nothing about the stability of the primal-dual recursion that seeks that optimum, the question addressed here.

Budget-pacing algorithms for online allocation drive the shadow price of budget by a mirror-descent update fed by observed spend \citep{balseiro2019,balseiro2020}, without a primal-side estimate distinct from that measurement, so the mismatch analyzed here does not arise. \citet{baelde2026companion} bound the oscillation of a Lagrangian reward signal at frozen policy, an $\mathcal O(\etalam/\omega)$ characterization, and defer the coupled regime this paper closes; dynamic-regret analyses of non-stationary CMDPs \citep{ding2023} bound performance against a drifting comparator with $\ghat = g$ throughout, whereas the mechanism here opens in a fixed periodic environment read through two different estimates.

\section{Setup}\label{sec:setup}

Two objects from the theory of linear systems carry the argument. The \emph{Floquet exponent} of Section~\ref{sec:parametric} plays for time-periodic coefficients the role the spectral radius plays for constant ones: the per-step logarithmic growth rate of the dominant solution, read off the spectral radius of the monodromy matrix over one period, whose sign decides stability. The \emph{top Lyapunov exponent} is the same growth rate for stochastic coefficients, the almost sure limit of $t^{-1}\log\|x_t\|$, and it is the object that survives once the mismatch broadens from a single line into a spectral density.

\subsection{Non-stationary constrained MDP}\label{sec:setup_cmdp}

Let $(\mathcal S, \mathcal A, P_\xi, R_\xi, C_\xi, \rho_0, \gamma)$ be a discounted MDP whose transition kernel, reward and cost depend on an exogenous process $\xi_t$ that carries the non-stationarity. Policies are parameterized by $\theta \in \mathbb R^d$, and for a frozen value $\xi$ of the exogenous process we write the return and the constraint functional
\begin{equation}\label{eq:functionals}
J(\pi_\theta;\xi) = \mathbb E_{\pi_\theta, P_\xi}\Big[\textstyle\sum_{t\ge0}\gamma^t R_\xi\Big], \qquad
g(\pi_\theta;\xi) = \mathbb E_{\pi_\theta, P_\xi}\Big[\textstyle\sum_{t\ge0}\gamma^t C_\xi\Big] - b,
\end{equation}
so that the constrained problem at frozen $\xi$ reads $\max_\theta J(\pi_\theta;\xi)$ subject to $g(\pi_\theta;\xi) \le 0$. The exogenous process enters both functionals.

\begin{assumption}[Slow exogenous variation]\label{ass:slow}
The exogenous process varies on a timescale long compared with the mixing time of the chain induced by any policy in the class, and its spectral content is concentrated at frequencies $\omega$ small compared with the inverse mixing time.
\end{assumption}

Assumption~\ref{ass:slow} lets \eqref{eq:functionals} describe the environment at time $t$ with $\xi = \xi_t$ (Appendix~\ref{sec:ec_setup_extra}). All frequencies in this paper are expressed in radians per learning step.

\begin{assumption}[Regular active saddle]\label{ass:saddle}
For a fixed reference value $\xi$ of the exogenous process, there exists $(\boldsymbol\theta^\star, \lambda^\star)$ with $\lambda^\star > 0$ and $g(\pi_{\boldsymbol\theta^\star};\xi) = 0$, the saddle of the frozen-$\xi$ problem \eqref{eq:functionals}, such that $J(\cdot;\xi)$ and $g(\cdot;\xi)$ are twice differentiable, $g(\cdot;\xi)$ has $\mathbf c := \partial g/\partial\boldsymbol\theta \neq 0$ in a neighbourhood of $\boldsymbol\theta^\star$, and the Lagrangian curvature $\mathbf A_L := -\partial^2 J/\partial\boldsymbol\theta^2 + \lambda^\star\,\partial^2 g/\partial\boldsymbol\theta^2 \succ 0$ at $\boldsymbol\theta^\star$: this is the second-order sufficient condition for $\boldsymbol\theta^\star$ to be a strict local saddle, stated here as a standing hypothesis; we do not derive it from properties of $J$ and $g$ alone.
\end{assumption}

From here on $\theta$, $a$ and $c$ denote the scalar reduction along $\mathbf c$, exact when $d=1$ or when $\mathbf c$ is an eigenvector of $\mathbf A_L$ and otherwise a modal approximation whose validity we do not establish, with the multidimensional case $K \ge 2$ treated in Remark~\ref{rem:multiK}. In the affine reduction used from Section~\ref{sec:setup_channels} on, $g$ is linear so $a$ reduces to the curvature of $J$ alone; a curved $g$ opens a distinct static channel through a biased estimate, which we do not pursue. The constraint is active at the saddle and the multiplier interior, so the projection in \eqref{eq:dual} is inactive in a neighbourhood and the tangent dynamics unconstrained. This is the regime of the whole analysis, and the operationally relevant one, since an inactive pacing constraint leaves the budget unspent. The active-projection regime is treated numerically, in Section~\ref{sec:exp_deeprl} and Appendix~\ref{sec:exp_prescription}. Section~\ref{sec:setup_channels} separates the two channels through which the exogenous process enters these functionals.

\subsection{The primal-dual learner, with two constraint signals}\label{sec:setup_dynamics}

The learner we study performs the projected updates
\begin{align}
\theta_{t+1} &= \theta_t + \etath\,\nabla_\theta\!\left[J(\pi_{\theta_t}) - \lambda_t\,\ghat(\pi_{\theta_t};\xi_t)\right],\label{eq:primal}\\
\lambda_{t+1} &= \proj{\lambda_t + \etalam\,g(\pi_{\theta_t};\xi_t)}.\label{eq:dual}
\end{align}

The structuring point of this paper is the distinction between $\ghat$, the constraint \emph{estimate} the primal update consumes inside its gradient, and $g$, the constraint \emph{measurement} feeding the dual update, two different objects in the implementations of Table~\ref{tab:deployed}. The primal needs a quantity differentiable in $\theta$, so it uses a learned cost critic, an analytic model, or a pathwise surrogate, consuming only its sensitivity $\partial\ghat/\partial\theta$. The dual needs no derivative, only a signed violation, so it uses the rawest available measurement, a realized spend or cost return. The two therefore differ in bias, in averaging window, and, decisively, in their response to a non-stationary environment: the critic, trained on recent data, lags a moving environment, while the raw counter reflects it without delay. The mismatch is a matter of degree, and the coherent case $\ghat = g$ is reachable in practice; Section~\ref{sec:algo} states what alignment requires.

\subsection{Ordering of the timescales as a choice}\label{sec:setup_timescales}

Let $\rratio := \etalam/\etath$ denote the timescale ratio. The two-timescale scheme of \citet[ch.~6]{borkar2008} is neutral about which variable is fast: what the theorem requires is a separation between its two step-size sequences (Appendix~\ref{sec:ec_setup_extra}). The assignment of the variables is a modeling decision. The CMDP instantiations are not neutral in practice, and they consistently make the multiplier the slowest variable, although \citet[§5]{borkar2005} notes that the latter two timescales could be interchanged. \citet{borkar2005} builds a constrained actor-critic on three timescales with the critic fastest, the actor intermediate and the dual ascent of the Lagrange multiplier slowest. \citet{bhatnagar2012} keeps that ordering, and \citet{panda2023} carries the same three-timescale construction to a finite-time analysis. The stated reason is exactly the quasi-static argument: a slow multiplier lets the actor-critic pair equilibrate against an essentially frozen price. So the analyzed regime is $\etalam \ll \etath$, that is $\rratio \ll 1$.

The regime in which the phenomenon of this paper lives is the opposite one, $\rratio \gg 1$: a dual step size that exceeds the primal one. Table~\ref{tab:deployed} reads the default step sizes out of three reference Lagrangian codebases; all three put the multiplier 2 orders of magnitude ahead of the policy, $\rratio \approx 10^2$, a choice with an operational motive, since a slow multiplier leaves a constraint violated for a long time. As $\rratio$ grows, the primal-dual pair leaves the overdamped side and acquires an oscillatory mode, a prerequisite for a parametric forcing to resonate with. Below a threshold on $\rratio$ the pair has no such mode, and the failure mode we analyze has no mechanism to act through. Sections~\ref{sec:additive} and~\ref{sec:parametric} make this precise: the damping of the pair, its worst-case amplification under additive forcing, and its tolerance to a sensitivity mismatch are all governed by $\rratio$, and all move against the practitioner as $\rratio$ grows. The gap we describe is therefore between an asymptotic theory and the practice that cites it.

The convergence guarantee that the CMDP literature invokes covers exactly the side on which the resonance cannot exist, so our result closes a coverage gap the convergence guarantee leaves open, without contradicting it. The corresponding prescription is stated in Section~\ref{sec:algo}.

\begin{table}
\TABLE
{Default step sizes in the reference Lagrangian implementations\label{tab:deployed}}
{\begin{tabular}{@{}lccc@{}}
\hline\up
implementation & policy & multiplier & $\rratio$ \\ \hline\up
\texttt{openai/safety-starter-agents}, PPO-Lagrangian & $3\times10^{-4}$ & $5\times10^{-2}$ & $167$ \\
\texttt{PKU-Alignment/omnisafe}, \texttt{PPOLag} & $3\times10^{-4}$ & $3.5\times10^{-2}$ & $117$ \\
\texttt{PKU-Alignment/Safe-Policy-Optimization}, single agent & $3\times10^{-4}$ & $3.5\times10^{-2}$ & $117$ \down\\ \hline
\end{tabular}}
{Defaults as of August 2026 (sources in Appendix~\ref{sec:ec_setup_extra}). The tabulated $\rratio$ is an upper bound, and all three sit on the $\rratio \gg 1$ side, past the boundary of Corollary~\ref{cor:existence}. In all three the primal consumes a learned cost critic and the dual a raw empirical cost, so $\ghat \ne g$ is the default.}
\end{table}

Constant step sizes and a fixed $\rratio$ make the sign of an exponent the available question, rather than a convergence statement (Appendix~\ref{sec:ec_setup_extra}).

\subsection{Two channels of non-stationarity}\label{sec:setup_channels}

\begin{definition}[Additive and parametric channels]\label{def:channels}
Let the exogenous process enter the constraint functional and the primal's estimate of it through their level and through their sensitivity, in the tangent coordinates of Assumption~\ref{ass:saddle},
\[
g(\pi_\theta;\xi_t) = c_{\lambda,t}\,\theta + b_t, \qquad
\ghat(\pi_\theta;\xi_t) = c_{\theta,t}\,\theta + b_t,
\]
so that $\partial\ghat/\partial\theta(\pi_\theta;\xi_t) = c_{\theta,t}$. The two signals share the same level term $b_t$ by construction, so no level bias between $\ghat$ and $g$ is modeled here. That choice is harmless for the exponents this paper studies, since a constant offset shifts the enforced level (Section~\ref{sec:algo}) without entering the linearized dynamics. The non-stationarity enters through the \emph{additive channel} when $b_t$ varies and $c_{\theta,t} = c_{\lambda,t} = c$ is constant, in which case the linearized learning dynamics are time-invariant and $b_t$ is an exogenous input. It enters through the \emph{parametric channel} when $c_{\theta,t}$ or $c_{\lambda,t}$ varies, in which case the coefficients of the linearized learning dynamics are themselves time-varying. Throughout, $b$ denotes the mean of $b_t$.
\end{definition}

An additive disturbance on a stable linear time-invariant system produces a response bounded by the frequency response, with no possibility of exponential growth; this is the implicit model behind treating environment drift as noise to average out (Section~\ref{sec:additive} confirms it for the constraint level). A parametric disturbance escapes that bound entirely, since a time-periodic coefficient can pump energy into an oscillatory mode at twice its natural frequency, growing exponentially for an arbitrarily small forcing amplitude once the pumping beats the damping. Section~\ref{sec:additive} therefore bounds a response, and Section~\ref{sec:parametric} tracks an exponent that can change sign.

The parametric channel is also the realistic one in the motivating application: over a day the pacing target moves, the additive channel, but what moves more is the marginal return of a unit of bid, since the same bid buys a different volume of impressions at 3am and at 9pm and converts at a different rate, a modulation of $\partial g/\partial\theta$ outside the reach of the standard analysis. Nor is the diurnal cycle an arbitrary forcing frequency: at the \texttt{omnisafe} step sizes of Table~\ref{tab:deployed}, the resonance $\omega = 2\wnat$ falls at a period of $N^\star \approx 970$ updates, the order of a day at usual update cadences (Appendix~\ref{sec:ec_setup_extra}).

\section{The linearized loop}\label{sec:additive}

Instantiate Assumptions~\ref{ass:slow} and \ref{ass:saddle} on the scalar linear-quadratic reduction that will serve throughout the paper: a concave quadratic return and an affine constraint forced sinusoidally in its level,
\begin{equation}\label{eq:toy}
J(\theta) = -\tfrac{a}{2}\theta^2, \qquad g(\theta;\xi_t) = c\,\theta + b + A\sin\omega t, \qquad \text{with } g \le 0,
\end{equation}
so that \eqref{eq:primal} and \eqref{eq:dual} become $\theta_{t+1} = \theta_t - \etath(a\theta_t + c\lambda_t)$ and $\lambda_{t+1} = \proj{\lambda_t + \etalam\,g(\theta_t;\xi_t)}$. The unforced stationary point is $\theta^\star = -b/c$ and $\lambda^\star = ab/c^2$, positive for $b > 0$, so the constraint is active and the projection stays inactive as long as the oscillation amplitude does not exceed $\lambda^\star$. In deviation coordinates the system is linear time-invariant with input $\etalam A\sin\omega t$ and step matrix
\begin{equation}\label{eq:stepmatrix_lti}
M = \begin{pmatrix} 1 - \etath a & -\etath c \\ \etalam c & 1\end{pmatrix}.
\end{equation}

Two structural facts about \eqref{eq:stepmatrix_lti} organize everything that follows. The characteristic polynomial is $z^2 - (2 - \etath a)z + (1 - \etath a + \etath\etalam c^2)$, whose value at $z = 1$ is $\etath\etalam c^2 > 0$ unconditionally, so the Jury criterion reduces to the single informative condition
\begin{equation}\label{eq:jury}
\etalam c^2 < a,
\end{equation}
the two side conditions holding with a wide margin throughout (Appendix~\ref{sec:ec_proof_additive}). The discriminant is $\etath^2a^2 - 4\etath\etalam c^2$. The poles are therefore complex, and the loop oscillatory, exactly when
\begin{equation}\label{eq:underdamped}
\rratio = \frac{\etalam}{\etath} > \frac{a^2}{4c^2}.
\end{equation}
Condition \eqref{eq:underdamped} is the regime of the deployed implementations of Table~\ref{tab:deployed} (Section~\ref{sec:setup_timescales}), and the regime in which the loop has a resonance to be excited. When the poles are complex we write $\sigma$ and $\wex$ for the decay rate and the frequency of the mode. They are available in closed form since both poles share the modulus $\sqrt{\det M}$,
\begin{equation}\label{eq:sigma_omega}
\sigma = -\tfrac12\log\big(1 - \etath a + \etath\etalam c^2\big), \qquad
\wex = \arg\Big(\tfrac{2-\etath a}{2} + \tfrac{i}{2}\sqrt{4\etath\etalam c^2 - \etath^2a^2}\Big),
\end{equation}
with the continuous-time approximations, valid to first order in the step sizes,
\begin{equation}\label{eq:wnat}
\wnat = c\sqrt{\etath\etalam}, \qquad \sigma \simeq \tfrac12\etath a, \qquad
\zeta := \frac{\sigma}{\wnat} \simeq \frac{a}{2c}\sqrt{\frac{\etath}{\etalam}} = \frac{a}{2c}\,\rratio^{-1/2}.
\end{equation}
The reduced damping $\zeta$ is the quantity that will appear in every threshold of this paper. It is normalized by $\wnat$, which is defined for every choice of step sizes, so that $\zeta$ remains meaningful on the overdamped side where $\wex$ does not exist. At the operating points where the loop oscillates, $\sigma/\wex$ differs from it at relative order $\zeta^2$. Interpreted directly, $\zeta \gg 1$ is the overdamped loop on which freezing the multiplier is licit, $\zeta \ll 1$ the oscillating loop on which it is not.

The transfer function from the constraint disturbance to the multiplier follows by eliminating the primal variable in the $z$ domain,
\begin{equation}\label{eq:transfer}
H_\lambda(z) = \frac{z - 1 + \etath a}{(z - 1 + \etath a)(z - 1) + \etath\etalam c^2},
\end{equation}
and the asymptotic peak-to-peak amplitude of the multiplier under the forcing of \eqref{eq:toy} is $\Delta\lambda = 2\etalam A\,|H_\lambda(e^{i\omega})|$. Setting $\etath = 0$ in \eqref{eq:transfer} recovers the frozen-policy case $H_\lambda(z) = 1/(z-1)$, hence $\Delta\lambda = \etalam A/|\sin(\omega/2)| \sim 2\etalam A/\omega$, which diverges as the forcing slows down and which reproduces the frozen-policy bound of \citet{baelde2026companion} including the $|\sin(\omega/2)|$ in the denominator.

\begin{proposition}[Additive channel: floor and worst case]\label{prop:floor}\label{prop:worstcase}
Under \eqref{eq:toy} and \eqref{eq:jury}:
\begin{enumerate}
\item[(i)] if $A < b$, so that the unprojected linearization stays valid throughout the cycle, the DC gain of \eqref{eq:transfer} is $H_\lambda(1) = a/(\etalam c^2)$ and $\lim_{\omega\to0}\Delta\lambda = 2Aa/c^2$, independent of both step sizes, with a plateau extending up to $\wnat\min\{1, (2\zeta)^{-1}\}$;
\item[(ii)] under \eqref{eq:underdamped} and to leading order in $\zeta$, $\sup_\omega \Delta\lambda \simeq A\etalam/\sigma$ with $2\sigma = \etath a - \etath\etalam c^2 + \mathcal O(\etath^2)$, that is $\sup_\omega \Delta\lambda \simeq (2Aa/c^2)\max\{1, (4\zeta^2)^{-1}\}$ with $(4\zeta^2)^{-1} = \rratio c^2/a^2$.
\end{enumerate}
\end{proposition}

The worst case over forcing frequencies is therefore amplified linearly by the timescale ratio. Both statements bear on the linearization, hence hold only while $A\etalam/\sigma < \lambda^\star$; past that point the projection truncates the excursion. Proofs and the discussion of the two branches are in Appendix~\ref{sec:ec_proof_additive}.

These two statements close the question left open at frozen policy by \citet{baelde2026companion}: the coupled low-frequency divergence does not exist, since the feedback through the policy adds a second pole and turns a pure integrator into a bandpass second-order system with finite DC gain. Both results concern a bounded amplitude; the rest of the paper treats the channel that admits exponential growth.

\section{The parametric channel}\label{sec:parametric}

Keep the reduction \eqref{eq:toy} but move the non-stationarity from the level to the sensitivity, and split the sensitivity into the two channels of Section~\ref{sec:setup_channels}. Let $s_t$ be a zero-mean normalized modulation signal and let
\begin{equation}\label{eq:modulation}
c_{\theta,t} = c\,(1 + \epsth s_t), \qquad c_{\lambda,t} = c\,(1 + \epslam s_t)
\end{equation}
be respectively the sensitivity seen by the primal update \eqref{eq:primal} inside its gradient and the sensitivity measured by the dual update \eqref{eq:dual} inside its violation signal. The recursion becomes $\theta_{t+1} = \theta_t - \etath(a\theta_t + c_{\theta,t}\lambda_t)$ and $\lambda_{t+1} = \proj{\lambda_t + \etalam(c_{\lambda,t}\theta_t + b)}$, with step matrix
\begin{equation}\label{eq:stepmatrix}
M_t = \begin{pmatrix} 1 - \etath a & -\etath c_{\theta,t} \\ \etalam c_{\lambda,t} & 1\end{pmatrix}.
\end{equation}
The offset terms dropped in passing to \eqref{eq:stepmatrix} are additive inputs, bounded by Proposition~\ref{prop:floor}, and cannot change the sign of an exponent (Appendix~\ref{sec:ec_param_extra}).

The coherent case is $\epsth = \epslam$, in which both sides of the loop see exactly the same non-stationarity.

Modulating the sensitivity seen by the primal modulates the \emph{damping} of the loop exactly as much as it modulates the \emph{stiffness}, through the same logarithmic derivative $\dot c_\theta/c_\theta$, whereas modulating the sensitivity measured by the dual modulates the stiffness alone. The two paths therefore contribute resonant forcing of opposite sign, and at the principal tongue they cancel. To see it, eliminate $\lambda$ from the continuous-time surrogate $\dot\theta = -\etath(a\theta + c_\theta\lambda)$, $\dot\lambda = \etalam c_\lambda\theta$ of \eqref{eq:stepmatrix}, which gives the second-order equation
\begin{equation}\label{eq:mathieu_raw}
\ddot\theta + \left(\etath a - \frac{\dot c_\theta}{c_\theta}\right)\dot\theta + \left(\etath\etalam c_\theta c_\lambda - \etath a\,\frac{\dot c_\theta}{c_\theta}\right)\theta = 0.
\end{equation}
Equation \eqref{eq:mathieu_raw} is a Hill equation, and for a sinusoidal modulation it is a Mathieu equation with damping.

Averaging \eqref{eq:mathieu_raw} at first order in the modulation depth produces a resonant forcing whose amplitude is
\begin{equation}\label{eq:bracket}
(\epsth + \epslam)\,\wnat^2 - \frac{\omega^2}{2}\,\epsth,
\end{equation}
with $\wnat^2 = \etath\etalam c^2$ as in \eqref{eq:wnat}. A parametric excitation pumps energy into an oscillatory mode most efficiently when its frequency is twice the natural frequency of that mode. The set of pairs (forcing frequency, modulation depth) at which the pumped mode has a positive exponent is the \emph{Arnold tongue} of the resonance. Without damping the tongue closes to a point on the frequency axis, so an arbitrarily small depth suffices; with damping it detaches from that axis and opens only above a threshold depth, which is the quantity Theorem~\ref{thm:threshold} computes. The tongue anchored at $\omega = 2\wex$ is the principal one, and it is the only one this paper analyzes. At $\omega = 2\wnat$ we have $\omega^2/2 = 2\wnat^2$ and the bracket \eqref{eq:bracket} collapses to $(\epslam - \epsth)\wnat^2$. The resonant forcing is proportional to the \emph{mismatch} between the two paths.

\begin{lemma}[Cancellation at the principal tongue]\label{lem:cancellation}
Let $s_t = \sin\omega t$ in \eqref{eq:modulation} and let $\theta$ obey \eqref{eq:mathieu_raw}. Take $\epsth, \epslam, \zeta$ of the same order, and expand to first order in that common small parameter. Then the slowly varying complex envelope $Z(t)$ of the resonant component $\theta(t) = \tfrac12\big(Z(t)e^{i\omega t/2} + \bar Z(t)e^{-i\omega t/2}\big)$ obeys
\[
\dot Z = -\sigma Z + \frac{\wnat}{4}\Big[(\epsth + \epslam) - \frac{\omega^2}{2\wnat^2}\epsth\Big]\bar Z,
\]
whose forcing coefficient reduces at $\omega = 2\wnat$ to $(\wnat/4)(\epslam - \epsth)$.
\end{lemma}

The proof is a first-order averaging of \eqref{eq:mathieu_raw} followed by a beat computation, given in full in Appendix~\ref{sec:ec_proofs}. Every later result rests on Lemma~\ref{lem:cancellation}: the cancellation is a structural protection, surviving beyond the sinusoidal case for the continuous flow and, under the condition of Remark~\ref{rem:secondorder}, for the recursion.

\begin{proposition}[Exact stability under coherent modulation]\label{thm:coherent}
Let $c_\theta(t) = c_\lambda(t) = c\,(1 + \varepsilon s(t))$ with $|\varepsilon| < 1$ and $|s(t)| \le 1$, so that $h = 0$ at every instant, and let $(\theta, \lambda)$ obey the continuous flow $\dot\theta = -\etath(a\theta + c_\theta\lambda)$, $\dot\lambda = \etalam c_\lambda\theta$ whose scalar reduction is \eqref{eq:mathieu_raw}. Then:
\begin{enumerate}
\item[(i)] the weighted energy $V = \etalam\theta^2 + \etath\lambda^2$ satisfies $\dot V = -2\etath\etalam a\,\theta^2 \le 0$ along every trajectory of the continuous flow, for every waveform $s$, every frequency and every depth. If $s$ is $T$-periodic, the monodromy matrix $\Phi$ over one period satisfies $\|\Phi\|_V < 1$, so the top Floquet exponent of the continuous flow is strictly negative. The recursion inherits this only under the condition of Remark~\ref{rem:secondorder};
\item[(ii)] for $s_t = \sin\omega t$ with $\omega = 2(\wnat + \Delta)$, to first order in $\zeta$ and to all orders in $\varepsilon$, the top Floquet exponent is
\[
\mu = -\sigma + \operatorname{Re}\sqrt{\sigma^2 J_1(\beta)^2 - \Delta^2}, \qquad \beta = \frac{2\wnat\varepsilon}{\omega},
\]
with $J_1$ the Bessel function of the first kind; at the principal tongue this is $-\sigma\big(1 - J_1(\varepsilon)\big)$, and the two Floquet multipliers remain a complex pair, with exponent exactly $-\sigma$, whenever $|\Delta| > \sigma J_1(\beta)$.
\end{enumerate}
\end{proposition}

Coherence removes the resonant parametric forcing of Lemma~\ref{lem:cancellation} identically and leaves a damping-borne pump bounded by $\sigma$ itself, smaller than the mismatched pump $\wnat|h|/4$ by a factor of order $\zeta$. Proofs and the extension to general periodic waveforms are in Appendices~\ref{sec:ec_param_extra} and~\ref{sec:ec_proof_coherent}.

\begin{remark}[The discrete recursion]\label{rem:secondorder}
The step matrix \eqref{eq:stepmatrix} inherits (ii) with a period-varying determinant and pump coefficient $\etath a/2$ in place of $\sigma$ (Appendix~\ref{sec:ec_param_extra}). Unlike (i), the result is not unconditionally negative: to first order in $\zeta$ it stays negative at every depth $\varepsilon \le 1$ if and only if $a\big(1 - J_1(1)\big) > \tfrac32\etalam c^2$, which every operating point in this paper satisfies with a wide margin.
\end{remark}

\begin{theorem}[Narrowband mismatch threshold]\label{thm:threshold}
Consider \eqref{eq:stepmatrix} with $s_t = \sin\omega t$ and $\omega$ in a neighbourhood of $2\wnat$ of width $\mathcal O(\zeta\wnat)$. To first order in the modulation depths and in $\zeta$, the leading Floquet exponent per step at $\omega = 2\wnat$ is
\begin{equation}\label{eq:floquet_rate}
\mu = \frac{|\epslam - \epsth|}{4}\,\wnat - \sigma,
\end{equation}
so, at the tongue centre and to the order of the expansion, the system is destabilized if and only if
\begin{equation}\label{eq:threshold}
|\epslam - \epsth| > 4\zeta, \qquad \zeta := \frac{\sigma}{\wnat},
\end{equation}
where $\zeta$ is exact given $\sigma$ of \eqref{eq:sigma_omega}, and to first order in $\etath\etalam c^2/a$,
\begin{equation}\label{eq:threshold_closed}
4\zeta = \frac{2a}{c}\sqrt{\frac{\etath}{\etalam}}\left(1 - \frac{\etalam c^2}{a}\right) + \mathcal O(\etath),
\end{equation}
the correction term being the same one Proposition~\ref{prop:worstcase} carries.
\end{theorem}

Condition \eqref{eq:threshold} is the classical damped-Mathieu threshold \citep{landau1976mechanics,nayfeh1979,rand2012}, here with the primal-dual mismatch $h = \epslam - \epsth$ playing the role of the modulation depth. What Lemma~\ref{lem:cancellation} supplies is that this specific difference of sensitivities is the effective depth, since the primal channel modulates damping and stiffness together while the dual channel modulates stiffness alone. The statement is narrowband; the broadband case is Proposition~\ref{prop:spectral}. Scope, the $\wnat$ versus $\wex$ distinction and the proof outline are in Appendix~\ref{sec:ec_param_extra}, the proof in Appendix~\ref{sec:ec_proof_threshold}.

\begin{corollary}[Existence condition at first order]\label{cor:existence}
Under the hypotheses of Theorem~\ref{thm:threshold}, the modulation depths of \eqref{eq:modulation} are bounded by $1$ on physical grounds, a sensitivity that reaches zero over the cycle being the extreme case, so $\sup|h| = 2$, not attained, and the criterion \eqref{eq:threshold} places the phenomenon behind a condition on the step sizes alone, independent of any amplitude,
\begin{equation}\label{eq:existence}
\rratio > \Big(\frac{a}{c}\Big)^2 .
\end{equation}
\end{corollary}

Only the overdamped half of this condition is exact; on the underdamped side it inherits the first order of \eqref{eq:threshold} and locates the boundary without bounding it: below $(a/c)^2/4$ it is structural by \eqref{eq:underdamped}, above it mildly optimistic (Appendices~\ref{sec:ec_param_extra} and~\ref{sec:ec_proof_existence}).

Condition \eqref{eq:existence} is an ordering condition on the algorithm's two learning rates alone, silent about the environment's magnitude, and the form the positioning of Section~\ref{sec:setup_timescales} takes.

\begin{proposition}[Existence of the exponent and its bandwidth scaling]\label{prop:spectral}
Let $s_t$ in \eqref{eq:modulation} be a zero-mean, stationary, ergodic signal of unit variance, uniformly bounded, whose spectral support is a band of width $B$ centred on $2\wex$, and let $\varepsilon$ denote the root-mean-square depth of the mismatch $\epslam - \epsth$. Then the top Lyapunov exponent of the random product of the matrices \eqref{eq:stepmatrix} exists and is almost surely constant, so the depth $\epscrit(B)$ at which it changes sign is well defined. It satisfies $\epscrit(0) = \varepsilon_0 := 2\sqrt2\,\zeta$, the narrowband threshold \eqref{eq:threshold} in root-mean-square convention, to the same first order in $\zeta$ as that threshold. For $B \gg \sigma$,
\begin{equation}\label{eq:spectral_asymptote}
\epscrit(B) = \varepsilon_0\sqrt{\frac{B}{\pi\sigma}}, \qquad B \gg \sigma,
\end{equation}
equivalently
\begin{equation}\label{eq:broadband}
\epscrit(B) \simeq \frac{2}{c}\sqrt{\frac{\kappa B a}{\etalam}}, \qquad B \gg \sigma.
\end{equation}
The one-parameter interpolation
\begin{equation}\label{eq:spectral}
\epscrit(B) = \varepsilon_0\sqrt{1 + \kappa\,\frac{B}{\sigma}}
\end{equation}
is exact at both ends, $B = 0$ above and $B \gg \sigma$ by \eqref{eq:spectral_asymptote}, and bridges them by construction.
\end{proposition}

The asymptote \eqref{eq:spectral_asymptote} follows from second-order stochastic averaging of the envelope equation of Lemma~\ref{lem:cancellation} against a broadband stationary mismatch, which draws on \citep{stratonovich1967,khasminskii1966,arnold1986,lin2004} (Appendix~\ref{sec:ec_param_extra}). The constant of the square-root law is $\kappa = 1/\pi \approx 0.3183$, set only by the profile of the tongue and independent of the step sizes, and measured in Appendix~\ref{sec:exp_broadband}. In \eqref{eq:broadband} the primal step size has cancelled: slowing the primal narrows the resonance tongue exactly as much as it reduces the damping.

Nothing in Lemma~\ref{lem:cancellation} requires the two modulations to be in phase: if $c_\theta$ and $c_\lambda$ share a waveform up to a phase offset $\Delta\varphi$, the threshold \eqref{eq:threshold} holds verbatim with $|\epslam - \epsth|$ replaced by $|\epslam - \epsth e^{i\Delta\varphi}|$, so a critic lagging the raw measurement (a worked instance, full derivation in Appendix~\ref{sec:ec_proofs}) is exactly the complex mismatch Theorem~\ref{thm:threshold} governs, not a benign approximation error. Several active constraints under a common waveform, and additive gradient noise, reduce exactly to the scalar case (Remark~\ref{rem:multiK}, Appendix~\ref{sec:ec_param_extra}).

\section{Algorithmic consequences}\label{sec:algo}

The primary prescription is to drive the primal and the dual from the same constraint estimate, $\ghat = g$. By Proposition~\ref{thm:coherent} this removes the resonant forcing by construction, at every modulation depth and bandwidth, with no new gain to tune and no step size altered. It leaves the unforced spectrum $(\sigma,\wnat)$ of \eqref{eq:stepmatrix_lti} exactly as it was, unlike every remedy below. What it forbids is the natural pattern of Section~\ref{sec:setup_channels}, the critic-versus-counter split that is exactly the mismatch of \eqref{eq:modulation}. Coherence means either differentiating through the same estimator that feeds the dual, or feeding the dual with the critic's own value in place of the raw measurement, which is the lighter change in most implementations. The dual then inherits the critic's bias, which has a consequence of its own that the analysis of this paper does not see. A dual update driven by $\ghat$ drives $\ghat$ to zero. A stationary bias $\delta$ in the critic then leaves the true constraint satisfied at the level $-\delta$, an offset of size $\delta$ from the nominal $0$ level. The offset is measurable against the raw counter and correctable by shifting the target, whereas the resonance is neither.

The secondary prescription acts on the step sizes and it is available in two strengths. The quantitative version is to bring $\etalam$ towards $\etath$, subject to the stability condition \eqref{eq:jury}: this raises $\zeta$, hence improves Proposition~\ref{prop:worstcase} and Theorem~\ref{thm:threshold} simultaneously, the two results coming from different channels that might have pulled in opposite directions. The ordered version is to push $\rratio$ below $(a/c)^2$, so that $4\zeta$ exceeds $\sup|h| = 2$: by Corollary~\ref{cor:existence} it acts on the existence of the failure mode, with no estimate of the modulation depth, and a design margin belongs below the structural bound $(a/c)^2/4$. A slower dual tracks a moving constraint more slowly, which is why practitioners do the opposite and the tradeoff \citet{stooke2020} document empirically; the analysis quantifies what is gained on the other side.

PID Lagrangian control \citep{stooke2020} and the classical augmented Lagrangian \citep{bertsekas1982} are competing remedies, and both cure the resonance the same way, by adding damping. The proportional gain adds $k_p\nabla g\nabla^\top\! g$ to the damping matrix of \eqref{eq:mathieu_raw}, positive semidefinite by construction, raising $\zeta$ and with it the threshold \eqref{eq:threshold}. The augmented multiplier update is dual ascent with the penalty coefficient $\rho$ in the role of $\etalam$, whose penalty term replaces the primal curvature $a$ by $a + \rho c^2$, a damping term of the same kind and already the mechanism of \citet{platt1988}. The two contributions of the penalty cancel in the determinant of \eqref{eq:stepmatrix}, so \eqref{eq:jury} holds for every $\rho$ and the tolerance \eqref{eq:threshold} exceeds that of the plain loop by the factor $(1 - \rho c^2/a)^{-1}$. A modulated sensitivity makes that cancellation inexact and opens a second, weak parametric channel of order $\etalam a\epsth/4$, by the same first-order argument as the curvature-modulation channel quantified in Appendix~\ref{sec:ec_controls}. Since $\rho$ is at once the penalty coefficient and the dual step size, the damping it supplies cannot be raised without a faster dual response, unlike a proportional gain. Both raise a threshold and leave the forcing in place, since neither says which constraint signal the primal update consumes. In the agent of Section~\ref{sec:exp_deeprl} the proportional gain doubles the linear threshold, from $0.15$ to $0.30$ at $\rratio = 100$, but that threshold lies where the additive forced response already drives the multiplier into its projection; there the PID agent locks in $2$ seeds out of $50$ at $\varepsilon = 0.4$ and $4$ out of $40$ at $\varepsilon \in \{0.3, 0.35\}$, against $29$ out of $50$ for the plain Lagrangian at $0.4$, while keeping the forced response and most of the return loss; the coherent agent halves that loss. Both remedies compose with $\ghat = g$, and the prescription is to do both.

Three connections to classical control practice follow. Anti-windup: the projection $\proj{\cdot}$ in \eqref{eq:dual} is a saturation on an integrator, and Appendix~\ref{sec:exp_prescription} shows it converts the divergence of Theorem~\ref{thm:threshold} into a limit cycle rather than removing it. Gain scheduling on the dual alone, by scheduling $\etalam$ on a known or predictable sensitivity, leaves $\epslam = 0$ with $\epsth$ untouched, hence a mismatch destabilizing whenever it exceeds $4\zeta$ (the row $(\epsth,\epslam) = (0.3,0)$ of Table~\ref{tab:floquet} is unstable). The schedule must therefore apply to both step sizes by the same profile, one way of realizing $\ghat = g$. The derivative term of \citet{stooke2020} amplifies whatever the primal and the dual disagree on, benign only under $\ghat = g$. Outside control practice, the reward-shaping alternative of \citet{baelde2026companion} sidesteps the question by removing the integrating dual variable, a tradeoff that Proposition~\ref{prop:floor} and Theorem~\ref{thm:threshold} now quantify.

\section{Numerical experiments}\label{sec:experiments}

The instantiation is the scalar reduction \eqref{eq:toy} and \eqref{eq:modulation}, which is the $K = 1$ case of the sinusoidal CMDP of \citet{baelde2026companion} with direct control of the forcing frequency, of the bandwidth and of the mismatch. We reuse that environment and keep its parameterization: $a = c = b = 1$ throughout, except in Section~\ref{sec:exp_deeprl}, forcing amplitude $A = 0.1$ in the additive experiments. Three instruments are used: closed-form frequency responses from \eqref{eq:transfer}, the Floquet exponent per step of the periodic monodromy, and the top Lyapunov exponent of the random matrix product. Appendix~\ref{sec:protocol} states their implementation and Appendix~\ref{sec:ec_controls} the control runs that validate each of them against a value known in advance. None of those controls measures the phenomenon itself. In the broadband experiments $\varepsilon$ denotes a root-mean-square depth, whereas in the periodic experiments it denotes a peak depth; for a sinusoid the peak is $\sqrt2$ times the root mean square.

The reference operating point of the parametric experiments, set by $\etath = 10^{-3}$ and $\etalam = 10^{-1}$ through \eqref{eq:sigma_omega}, is reported in Table~\ref{tab:opoint}. The exponents and thresholds measured below are swept in modulation period, so they are quoted against the discrete threshold $4\sigma/\wex = 0.18022$ (peak) and $2\sqrt2\,\sigma/\wex = 0.12744$ (root-mean-square), built on the exact discrete frequency $\wex$. Its relative gap to the continuous approximation $\wnat$, $8\times10^{-4}$, is small against the tongue's width but large against the precision reported below, which is why $\wex$ is carried through this section. The additive channel behaves as Proposition~\ref{prop:floor} predicts, with the tracking plateau and corner recovered exactly and the worst case within a factor $1.00$ to $1.47$ (Appendix~\ref{sec:exp_additive}).

\subsection{The parametric channel under narrowband forcing}\label{sec:exp_narrowband}

\begin{figure}
\FIGURE
{\includegraphics{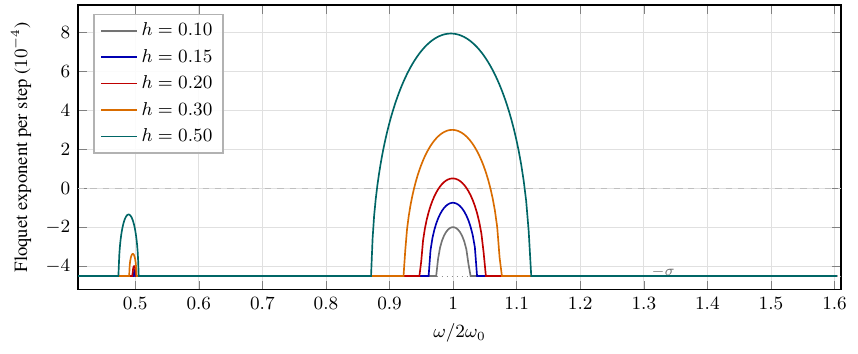}}
{Measured resonance tongue: Floquet exponent per step against the forcing frequency\label{fig:tongue}}
{Operating point $\etath = 10^{-3}$, $\etalam = 10^{-1}$ of Table~\ref{tab:floquet}, pure dual modulation $\epsth = 0$, $\epslam = h \in \{0.10,0.15,0.20,0.30,0.50\}$, exponent read off the monodromy over one exact period. The dotted line is the unmodulated exponent $-\sigma = -4.502\times10^{-4}$, reproduced outside the two tongues; the threshold $4\sigma/\wex = 0.18022$ separates $h=0.15$ from $h=0.20$.}
\end{figure}

The failure mode is confirmed specific to the coupled system and localized to the tongue: it vanishes at frozen policy, matches the closed form at zero depth, is confined to $2\wex$ at fixed mismatch, and disappears when the modulation moves from the sensitivity to the curvature. Figure~\ref{fig:tongue} shows a single narrow instability region around $2\wex$, flat elsewhere, whose peak tilts downward in frequency as the depth grows; a secondary tongue at $\omega/2\wex = 1/2$ degrades the damping without reversing it over the depths swept here; Table~\ref{tab:floquet} confirms the sign is still unchanged at $h = 0.9$.

Table~\ref{tab:floquet} tests the closed-form exponent \eqref{eq:floquet_rate}, a sharper test than the threshold alone: the fitted-quantity-free prediction $\mu = |h|\wex/4 - \sigma$ is accurate to about $1\%$ up to $|h| = 0.3$ and to $2.4\%$ at $|h| = 0.9$, reproduces the sign change bracketing the threshold, and confirms that only the mismatch $h = \epslam - \epsth$ matters.

\begin{table}
\TABLE
{Parametric channel: predicted against measured Floquet exponent per step\label{tab:floquet}}
{\begin{tabular}{@{}lccrr@{}}
\hline\up
$\epsth$ & $\epslam$ & $h = \epslam - \epsth$ & predicted $\mu$ & measured $\mu$ \\ \hline\up
0 & 0 & 0 & $-4.502\times10^{-4}$ & $-4.502\times10^{-4}$ \\
0 & 0.15 & 0.15 & $-7.55\times10^{-5}$ & $-7.52\times10^{-5}$ \\
0 & 0.20 & 0.20 & $+4.94\times10^{-5}$ & $+4.97\times10^{-5}$ \\
0.3 & 0 & $-0.30$ & $+2.992\times10^{-4}$ & $+2.988\times10^{-4}$ \\
0.2 & 0.5 & 0.30 & $+2.992\times10^{-4}$ & $+2.961\times10^{-4}$ \\
0 & 0.90 & 0.90 & $+1.798\times10^{-3}$ & $+1.755\times10^{-3}$ \\
0.9 & 0.9 & 0 & $-2.276\times10^{-4}$ & $-2.271\times10^{-4}$ \down\\ \hline
\end{tabular}}
{Operating point $\etath = 10^{-3}$, $\etalam = 10^{-1}$, forcing at $\omega = 2\wex$. Predicted $\mu = |h|\wex/4 - \sigma$ from \eqref{eq:floquet_rate}, with $\wex = 0.00999216$ and $\sigma = 4.502\times10^{-4}$ from \eqref{eq:sigma_omega}. The last row is a coherent modulation, predicted by Remark~\ref{rem:secondorder}. Omitted rows: $(\epsth,\epslam) = (0, 0.3)$ measures $+2.988\times10^{-4}$, and the coherent $(0.3, 0.3)$ measures $-3.748\times10^{-4}$ against $-3.750\times10^{-4}$.}
\end{table}

The last row of Table~\ref{tab:floquet} probes the discrete exponent of Remark~\ref{rem:secondorder}: a coherent modulation degrades the damping without reversing its sign, matching the prediction to $0.25\%$ at depths $0.3$ and $0.9$. At depth $0.9$, a coherent modulation gives $-2.271\times10^{-4}$ while a dual-only modulation gives $+1.755\times10^{-3}$, the contrast Proposition~\ref{thm:coherent} and Theorem~\ref{thm:threshold} predict.

Table~\ref{tab:ec_rsweep} tests the collapse of the tolerance across 4 decades of timescale ratio, matching $4\zeta$ to $1\%$ where $\zeta$ is small. At the boundary case $\rratio = 1$, where the first-order criterion is undecided, the exact map resolves it below the admissible bound.

Table~\ref{tab:borkar} approaches the same boundary from the other side and materializes the coverage gap of Section~\ref{sec:setup_timescales}, sweeping the timescale ratio at fixed $\etath\etalam = 10^{-4}$ with the mismatch distributed symmetrically at its largest admissible value, $\epsth = +0.99$ and $\epslam = -0.99$.

\begin{table}
\TABLE
{Existence boundary of the phenomenon: mismatch threshold maximized over the forcing frequency\label{tab:borkar}}
{\begin{tabular}{@{}lccccc@{}}
\hline\up
$\rratio = \etalam/\etath$ & $\wex$ & $\zeta$ & measured threshold & $4\sigma/\wex$ & $4\sigma/\wnat$ \\ \hline\up
$10^{-1}$ & real spectrum & $>1$ & none ($\le 1.98$) & n/a & n/a \\
$0.25$ & real spectrum & $>1$ & none ($\le 1.98$) & n/a & n/a \\
$0.96$ & 0.008644 & 0.5079 & none ($\le 1.98$) & 2.3503 & 2.0315 \\
$0.98$ & 0.008674 & 0.5026 & 1.9673 & 2.3176 & 2.0104 \\
$1$ & 0.008704 & 0.4975 & 1.9484 & 2.2863 & 1.9899 \\
$10^{2}$ & 0.009992 & 0.045020 & 0.1802 & 0.18022 & 0.18008 \down\\ \hline
\end{tabular}}
{Fixed $\etath\etalam = 10^{-4}$ so that $\wnat = 10^{-2}$ is constant, $a=c=1$, mismatch distributed symmetrically, exponent maximized over modulation periods from $8$ to $3000$ steps. The last two columns are the two forms of the first-order threshold \eqref{eq:threshold}, $4\sigma/\wex$ (discrete frequency) and $4\sigma/\wnat=4\zeta$ (continuous surrogate), differing at relative order $\zeta^2$ hence $15\%$ apart near $\rratio=1$ and $0.08\%$ at $\rratio=10^2$. ``None'' means the exponent stays negative at every period swept; the row $\rratio=0.25$ is the exact overdamping boundary $(a/c)^2/4$ of \eqref{eq:underdamped}. Omitted rows: $\rratio = 10^{-2}$ and $0.85$ give none; $\rratio = 10$ gives $0.6125$ against $0.6202$ and $0.6134$.}
\end{table}

In the timescale ordering the CMDP literature prescribes, $\rratio \le 0.1$, the spectrum is real and the mechanism is structurally absent, up to $\rratio = (a/c)^2/4 = 0.25$ exactly where \eqref{eq:underdamped} places the transition. The existence boundary sits slightly below the first-order prediction \eqref{eq:existence}, resolved to about $1\%$ against the prediction's own $\mathcal O(\zeta^2)$ ambiguity of $15\%$, consistent with Corollary~\ref{cor:existence}. Just past the crossing the growth is slow: doubling times of $20\,600$ and $8\,300$ steps just past the crossing at $\rratio = 0.98$ and $1$, against $152$ steps at $\rratio = 10^{2}$, with the agreement degrading as $\zeta$ grows towards $1$ (Table~\ref{tab:borkar_refine} refines the crossing bracket).

\subsection{Broadband forcing and the prescription}\label{sec:exp_summary}

Under broadband forcing the threshold follows \eqref{eq:spectral} over a factor $64$ in bandwidth, with $\kappa = 0.309 \pm 0.029$ measured against $1/\pi \approx 0.318$, and the primal step size drops out at fixed physical bandwidth (Appendix~\ref{sec:exp_broadband}, Figure~\ref{fig:bandwidth}). At equal modulation depth a coherent signal contracts over the horizon while a dual-only mismatch grows, 6 orders of magnitude apart, and with the projection active the instability persists as a bounded limit cycle (Appendix~\ref{sec:exp_prescription}).

\subsection{A deep reinforcement learning agent}\label{sec:exp_deeprl}

The reduction predicts the failure of an implementation pattern, so we test it on an agent that has the pattern without being built for it. The environment is a one-step constrained bandit with action $\theta \in [0,1]$, reward $\theta - \theta^2/2$ and cost $c_t\theta - b$, $c_t = 1 + \varepsilon\sin\omega t$, $b = 0.3$, whose phase the agent does not observe; at the saddle its curvature and sensitivity are those of \eqref{eq:modulation} with $a = c = 1$; the linear term of the reward places the saddle at $(\theta^\star, \lambda^\star) = (b, 1 - b) = (0.3, 0.7)$. The learner is Lagrangian SAC with a reward critic $Q_r$ and a cost critic $Q_c$ fit on a replay buffer of $5\,000$ transitions, an actor trained by SGD against $Q_r - \lambda Q_c$, and the projected dual ascent \eqref{eq:dual} on the cost averaged over the last $W = 10$ steps. The actor uses plain SGD so that its gains are constant and can be identified; adaptive optimizers such as Adam rescale the step by running gradient statistics, which this test does not cover. The buffer spans many modulation periods, so the actor's constraint estimate is unmodulated while the dual's measurement is modulated: $\epsth \approx 0$ and $\epslam = \varepsilon$, the critic-versus-counter split of Section~\ref{sec:setup_channels}. Two variants change one element each. The \emph{coherent} agent feeds $(\sin\omega t, \cos\omega t)$ to $Q_c$ and queries it at the current phase, so that both updates see the modulated sensitivity, $h \approx 0$, with the dual path, its noise and its delay unchanged. The \emph{PID} agent adds a proportional term, $\lambda = \proj{I + k_p g}$ with $k_p = 1$ and $I$ the integral of \eqref{eq:dual} \citep{stooke2020}. The response is the amplitude $A_{1/2}$ of $\lambda$ at $\omega/2$, demodulated over the second half of a $20\,000$-step run; a run is \emph{locked} when $A_{1/2} > 0.4$, against $0.12$ for unmodulated runs.

The linearized loop of the agent has no closed form, so it is identified. On unmodulated runs, a least-squares fit of $\delta\theta_t = (1-k_a)\,\delta\theta_{t-1} - k_\lambda\,\delta\lambda_t$ on the deterministic action gives the effective primal gains, which play the roles of $\etath a$ and $\etath c$, and the Floquet exponent of the identified loop, with the $W$-step dual window kept exactly, gives a predicted threshold $\varepsilon_c$ with no fitted modulation parameter. At the actor step size $3\times10^{-3}$ and $\etalam = 0.1$, the fit gives $k_a = 1.74\times10^{-3}$ and $k_\lambda = 1.58\times10^{-3}$, hence an effective ratio $\rratio = \etalam a/k_a = 58$ and $\wex = 0.0126$. The averaging window delays the dual and halves the reduced damping, from $0.063$ at $W = 1$ to $0.034$, and the exact Floquet exponent of the identified loop predicts $\varepsilon_c = 0.19$, above the first-order value $4\zeta = 0.14$ of Theorem~\ref{thm:threshold}, at $\omega = 0.026 \approx 2\wex$. The measured lock-in sets in between $0.15$ and $0.3$, with $1$, $4$ and $10$ seeds out of $10$ locked at $\varepsilon = 0.15$, $0.2$ and $0.3$. Holding $\lambda$ at its saddle value for the first $5\,000$ steps gives $2$, $4$ and $8$ locked seeds, so the lock-in does not depend on the start-up transient.

The timescale ratio is then varied at fixed $\etath\etalam$, hence fixed $\wex$, by scaling the actor and dual step sizes to $\rratio \in \{1, 10, 100\}$; at $\rratio = 100$ they are $2.28\times10^{-3}$ and $\etalam = 0.132$, with identified gains $k_a = 1.324\times10^{-3}$ and $k_\lambda = 1.204\times10^{-3}$, and Appendix~\ref{sec:protocol} lists them at every ratio. The gains identified on unmodulated runs give effective ratios $1.1$, $10.2$ and $99.5$ and predicted thresholds at $\omega = 0.026$ of $1.85$, $0.53$ and $0.15$ for the Lagrangian agent and of none, $1.18$ and $0.30$ for the PID agent. The $\sqrt{\rratio}$ law of Theorem~\ref{thm:threshold} therefore predicts that at $\varepsilon = 0.4$ the Lagrangian agent locks at $\rratio = 100$ and stays unlocked at $\rratio = 10$, that at $\varepsilon = 0.8$ it locks at $\rratio = 10$ and stays unlocked at $\rratio = 1$, that the PID agent still locks at $\rratio = 100$, $\varepsilon = 0.4$, and that the coherent agent locks nowhere. These four predictions were fixed before the runs, those at $\varepsilon = 0.8$ revised before their data were read.

\begin{figure}
\FIGURE
{\includegraphics{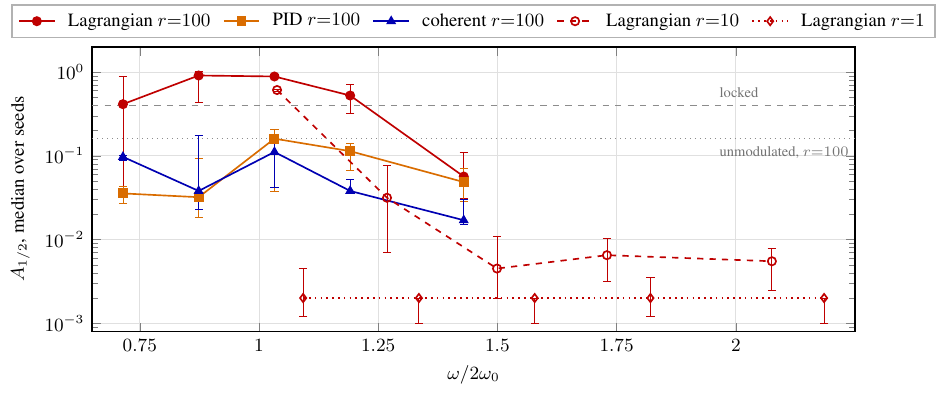}}
{Lagrangian SAC: sub-harmonic amplitude of the multiplier against the forcing frequency\label{fig:deeprl}}
{Median of $A_{1/2}$ over $10$ seeds ($5$ for the coherent agent), bars from the first to the third quartile. Each frequency is divided by the $2\wex$ of its own identified loop, $0.0252$, $0.0173$ and $0.0165$ at $\rratio = 100$, $10$ and $1$. The $\rratio = 100$ curves are at $\varepsilon = 0.4$, $b = 0.3$; the $\rratio = 10$ and $\rratio = 1$ curves at $\varepsilon = 0.8$, $b = 0.15$. Dashed line: lock criterion $0.4$; dotted line: unmodulated level at $\rratio = 100$, $0.159$.}
\end{figure}

Figure~\ref{fig:deeprl} gives the outcome, with each frequency normalized by the $2\wex$ of its own identified loop. The Lagrangian agent locks in $10$ seeds out of $10$ at $\rratio = 100$, $\varepsilon = 0.4$, $\omega = 0.026$, and in none of $10$ at $\rratio = 10$. The cells at $\varepsilon = 0.8$ run at $b = 0.15$, since the instantaneous saddle $b/c_t$ stays inside the action interval only for $b < 1 - \varepsilon$, with the gains identified again and the predictions revised before the runs were read (Appendix~\ref{sec:ec_limits}): at $\rratio = 10$ the loop is unstable only at $\omega = 0.018$, and at $\rratio = 1$ it is stable everywhere. The Lagrangian agent locks in $9$ seeds out of $10$ at $\rratio = 10$, $\omega = 0.018$, in none at the four other frequencies, and in none of the $50$ runs at $\rratio = 1$, so the $\sqrt{\rratio}$ law holds on both comparisons, each at a single frequency. The coherent agent locks in none of its $75$ runs, $15$ cells of $5$ seeds, including the $5$ cells at $\rratio = 100$ where the Lagrangian agent locks in $29$ seeds out of $50$.

The PID prediction fails. At $\rratio = 100$, $\varepsilon = 0.4$ the PID agent locks in $2$ seeds out of $50$, and in none of $10$ at $\omega = 0.026$, where at least $6$ were predicted. Neither the gains nor the horizon account for it: identified on the PID agent's own unmodulated runs, the gains match the Lagrangian ones within $1\%$ and give a threshold of $0.29$, and the Floquet exponent at $\omega = 0.026$ is $+3.3\times10^{-4}$ per step, a growth factor of about $5\times10^2$ over the $19\,000$ learning steps. The cause is the term that Lemma~\ref{lem:cancellation} sets aside by working in deviation coordinates (Appendix~\ref{sec:ec_param_extra}). Modulating the cost also feeds the dual an additive input $\etalam\varepsilon\bar\theta\sin\omega t$, with $\bar\theta = 0.291$ the mean action, whose forced response has amplitude $A_\omega = \etalam\varepsilon\bar\theta\,\omega/|\omega^2 - \wex^2|$ when $k_a \ll \omega$. At $\omega = 0.026$, $\varepsilon = 0.4$ it is $0.77$, above $\lambda^\star = 0.70$: the trough of the forced oscillation reaches the projection, which holds $\lambda$ at zero $25\%$ of the time in the Lagrangian runs and clips $A_\omega$ to the measured $0.63$. The proportional gain does double the linear threshold, from $0.15$ to $0.30$, but that threshold lies where the forced response already reaches the projection, and the saturation, not the linear damping, decides the outcome. A confirmation lot at $\varepsilon \in \{0.3, 0.35\}$ gives $2$ locked seeds out of $20$ at $\omega = 0.026$, against $9$ for the identified loop simulated with its projection and $16$ for the linear model, both fixed before the runs: the agent saturates more than the simulated loop, and suppresses more (Appendix~\ref{sec:exp_prescription}). The same saturation widens the Lagrangian tongue: the agent locks at $\omega = 0.018$, $0.022$ and $0.030$ ($5$, $7$ and $6$ seeds) where the linear exponent is negative.

Locking lowers the return and leaves the constraint satisfied (Table~\ref{tab:deeprl_cost}). At $\rratio = 100$, $\varepsilon = 0.4$ the mean windowed constraint value is negative in every cell, between $-0.040$ and $-0.002$, and the return of the deterministic action falls from $0.248$ unmodulated to $0.227$ for the Lagrangian agent at $\omega = 0.026$. The coherent agent halves that loss at the tongue centre, $0.238$ against $0.232$ for the PID agent, and has the best return on $\omega \in [0.022, 0.030]$ only: at $\omega = 0.018$ and $0.036$ the PID agent does slightly better. There the coherent agent's $A_{1/2}$ stays below the unmodulated level, so its loss goes with the additive forced response, which a coherent signal leaves in place. The PID agent, which removes the sub-harmonic and keeps the forced response, recovers on the tongue between a fifth and a quarter of the Lagrangian loss.

\section{Conclusion}\label{sec:conclusion}

Separating the constraint estimate consumed by the primal update from the constraint measurement consumed by the dual update opens a second channel of non-stationarity, one that modulates the coefficients of the linearized recursion and whose only driver is the sensitivity mismatch $\epslam - \epsth$ (Lemma~\ref{lem:cancellation}). The resonance this opens sits where the practice sits, outside the region the theory was built to secure: at the $\rratio \approx 10^2$ of Table~\ref{tab:deployed}, the tolerance is an order of magnitude below its value at $\rratio = 1$, opposite the $\rratio \ll 1$ regime of \citet{borkar2008,borkar2005,bhatnagar2012}.

The prescription needs no new machinery: drive both updates from the same constraint signal (Proposition~\ref{thm:coherent}), and where that is unavailable, move $\etalam$ towards $\etath$ (Theorem~\ref{thm:threshold}, Corollary~\ref{cor:existence}), verified numerically on the affine recursion (Section~\ref{sec:experiments}). On a Lagrangian soft actor-critic agent the coherent critic removes the lock-in in every cell, the $\sqrt{\rratio}$ law holds at the two ratios compared, and the projection of the multiplier, outside the linear theory, widens the tongue and decides the outcome of a PID variant (Section~\ref{sec:exp_deeprl}). The statements are proved on the tangent linear-quadratic reduction of Section~\ref{sec:setup_cmdp}, whose scope the next paragraph delimits. By Remark~\ref{rem:multiK} the statements extend exactly to $K$ constraints under a common waveform.

\label{sec:scope}

Two extensions remain open. The first is a multidimensional policy parameter: the scalar pair $(a, c)$ is exact only when $d = 1$ or when the constraint gradient is an eigenvector of the curvature, and otherwise the tongue comes from a coupled system of dimension $d + 1$. We expect the mechanism to persist on the mode most strongly coupled to the constraint, but reducing a $d$-dimensional curvature to the single number $a$ carries an uncontrolled error; closing that gap is a spectral question about the pencil, not a new dynamical mechanism. The second is constraints driven by \emph{distinct} waveforms. Remark~\ref{rem:multiK} settles the common-waveform case exactly under its isotropy condition on the curvature; distinct frequencies instead open several tongues that can cross as the operating point moves, and nothing here says what happens at an intersection. This is left for future work; it is not addressed by \citet{baelde2026companion}, whose reward-shaping construction removes the dual variable and does not analyze its resonance structure.

A separate, unresolved puzzle: \citet[Figure 4]{stooke2020} sweep the integral gain $K_I$, our $\etalam$, and observe the cost oscillations shrink and vanish as the gain grows, which read naively runs against Proposition~\ref{prop:worstcase} and Theorem~\ref{thm:threshold}. We have not run their sweep ourselves. A conjectured reconciliation: their perturbation has a spectrum fixed by the rate of policy improvement rather than swept, so as $\etalam$ grows $\wnat \propto \sqrt{\etalam}$ leaves that excitation band, predicting an amplitude non-monotone in $K_I$, maximal where $\wnat$ crosses it. That prediction is testable on the same sweep, which we leave to future work.
\paragraph{Declaration of generative AI and AI-assisted technologies in the manuscript preparation process.} During the preparation of this work the author used Claude (Anthropic) in order to assist with drafting and editing the text and with writing the numerical scripts. After using this tool, the author reviewed and edited the content as needed and takes full responsibility for the content of the article.

\bibliography{references}

\begin{thebibliography}{36}
\providecommand{\natexlab}[1]{#1}
\providecommand{\url}[1]{\texttt{#1}}
\expandafter\ifx\csname urlstyle\endcsname\relax
  \providecommand{\doi}[1]{doi: #1}\else
  \providecommand{\doi}{doi: \begingroup \urlstyle{rm}\Url}\fi

\bibitem[Achiam et~al.(2017)Achiam, Held, Tamar, and Abbeel]{achiam2017}
Joshua Achiam, David Held, Aviv Tamar, and Pieter Abbeel.
\newblock Constrained policy optimization.
\newblock In \emph{Proceedings of the 34th International Conference on Machine
  Learning}, volume~70 of \emph{Proceedings of Machine Learning Research}, pp.\
   22--31, 2017.

\bibitem[Altman(1999)]{altman1999}
Eitan Altman.
\newblock \emph{Constrained {M}arkov Decision Processes}.
\newblock Chapman \& Hall/CRC, 1999.

\bibitem[Arnold et~al.(1986)Arnold, Papanicolaou, and Wihstutz]{arnold1986}
Ludwig Arnold, George Papanicolaou, and Volker Wihstutz.
\newblock Asymptotic analysis of the {L}yapunov exponent and rotation number of
  the random oscillator and applications.
\newblock \emph{SIAM Journal on Applied Mathematics}, 46\penalty0 (3):\penalty0
  427--450, 1986.

\bibitem[Arrow et~al.(1958)Arrow, Hurwicz, and Uzawa]{arrow1958}
Kenneth~J. Arrow, Leonid Hurwicz, and Hirofumi Uzawa.
\newblock \emph{Studies in Linear and Non-Linear Programming}.
\newblock Stanford Mathematical Studies in the Social Sciences. Stanford
  University Press, 1958.

\bibitem[Baelde(2026)]{baelde2026companion}
Maxime Baelde.
\newblock Companion paper on constrained reinforcement learning.
\newblock Under review, 2026.

\bibitem[Balseiro et~al.(2020)Balseiro, Lu, and Mirrokni]{balseiro2020}
Santiago Balseiro, Haihao Lu, and Vahab Mirrokni.
\newblock Dual mirror descent for online allocation problems.
\newblock In \emph{Proceedings of the 37th International Conference on Machine
  Learning}, volume 119 of \emph{Proceedings of Machine Learning Research},
  pp.\  613--628, 2020.

\bibitem[Balseiro \& Gur(2019)Balseiro and Gur]{balseiro2019}
Santiago~R. Balseiro and Yonatan Gur.
\newblock Learning in repeated auctions with budgets: Regret minimization and
  equilibrium.
\newblock \emph{Management Science}, 65\penalty0 (9):\penalty0 3952--3968,
  2019.

\bibitem[Benveniste et~al.(1990)Benveniste, M{\'e}tivier, and
  Priouret]{benveniste1990}
Albert Benveniste, Michel M{\'e}tivier, and Pierre Priouret.
\newblock \emph{Adaptive Algorithms and Stochastic Approximations}, volume~22
  of \emph{Stochastic Modelling and Applied Probability}.
\newblock Springer-Verlag, 1990.

\bibitem[Bernstein et~al.(2019)Bernstein, Dall'Anese, and
  Simonetto]{bernstein2019}
Andrey Bernstein, Emiliano Dall'Anese, and Andrea Simonetto.
\newblock Online primal-dual methods with measurement feedback for time-varying
  convex optimization.
\newblock \emph{IEEE Transactions on Signal Processing}, 67\penalty0
  (8):\penalty0 1978--1991, 2019.

\bibitem[Bertsekas(1982)]{bertsekas1982}
Dimitri~P. Bertsekas.
\newblock \emph{Constrained Optimization and Lagrange Multiplier Methods}.
\newblock Academic Press, 1982.

\bibitem[Bhatnagar \& Lakshmanan(2012)Bhatnagar and Lakshmanan]{bhatnagar2012}
Shalabh Bhatnagar and K.~Lakshmanan.
\newblock An online actor-critic algorithm with function approximation for
  constrained {M}arkov decision processes.
\newblock \emph{Journal of Optimization Theory and Applications}, 153\penalty0
  (3):\penalty0 688--708, 2012.

\bibitem[Borkar(2005)]{borkar2005}
Vivek~S. Borkar.
\newblock An actor-critic algorithm for constrained {M}arkov decision
  processes.
\newblock \emph{Systems \& Control Letters}, 54\penalty0 (3):\penalty0
  207--213, 2005.

\bibitem[Borkar(2008)]{borkar2008}
Vivek~S. Borkar.
\newblock \emph{Stochastic Approximation: A Dynamical Systems Viewpoint}.
\newblock Cambridge University Press, 2008.
\newblock Chapter 6, two-timescale stochastic approximation.

\bibitem[Colombino et~al.(2020)Colombino, Dall'Anese, and
  Bernstein]{colombino2020}
Marcello Colombino, Emiliano Dall'Anese, and Andrey Bernstein.
\newblock Online optimization as a feedback controller: Stability and tracking.
\newblock \emph{IEEE Transactions on Control of Network Systems}, 7\penalty0
  (1):\penalty0 422--432, 2020.

\bibitem[Ding \& Lavaei(2023)Ding and Lavaei]{ding2023}
Yuhao Ding and Javad Lavaei.
\newblock Provably efficient primal-dual reinforcement learning for {CMDP}s
  with non-stationary objectives and constraints.
\newblock In \emph{Proceedings of the AAAI Conference on Artificial
  Intelligence}, volume~37, pp.\  7396--7404, 2023.
\newblock \doi{10.1609/aaai.v37i6.25900}.
\newblock arXiv:2201.11965.

\bibitem[Feijer \& Paganini(2010)Feijer and Paganini]{feijer2010}
Diego Feijer and Fernando Paganini.
\newblock Stability of primal-dual gradient dynamics and applications to
  network optimization.
\newblock \emph{Automatica}, 46\penalty0 (12):\penalty0 1974--1981, 2010.

\bibitem[Hauswirth et~al.(2024)Hauswirth, He, Bolognani, Hug, and
  D{\"o}rfler]{hauswirth2024}
Adrian Hauswirth, Zhiyu He, Saverio Bolognani, Gabriela Hug, and Florian
  D{\"o}rfler.
\newblock Optimization algorithms as robust feedback controllers.
\newblock \emph{Annual Reviews in Control}, 57:\penalty0 100941, 2024.
\newblock Preprint arXiv:2103.11329v2, 2024.

\bibitem[Ji et~al.(2024)Ji, Zhou, Zhang, Dai, Pan, Sun, Huang, Geng, Liu, and
  Yang]{omnisafe2024}
Jiaming Ji, Jiayi Zhou, Borong Zhang, Juntao Dai, Xuehai Pan, Ruiyang Sun,
  Weidong Huang, Yiran Geng, Mickel Liu, and Yaodong Yang.
\newblock {O}mni{S}afe: An infrastructure for accelerating safe reinforcement
  learning research.
\newblock \emph{Journal of Machine Learning Research}, 25\penalty0
  (285):\penalty0 1--6, 2024.

\bibitem[Khalil(2002)]{khalil2002}
Hassan~K. Khalil.
\newblock \emph{Nonlinear Systems}.
\newblock Prentice Hall, 3 edition, 2002.
\newblock Chapter 9, Stability of Perturbed Systems.

\bibitem[Khasminskii(1966)]{khasminskii1966}
R.~Z. Khasminskii.
\newblock A limit theorem for solutions of differential equations with random
  right-hand sides.
\newblock \emph{Theory of Probability \& Its Applications}, 11\penalty0
  (3):\penalty0 390--406, 1966.

\bibitem[Krsti{\'c} \& Wang(2000)Krsti{\'c} and Wang]{krstic2000}
Miroslav Krsti{\'c} and H.-H. Wang.
\newblock Stability of extremum seeking feedback for general nonlinear dynamic
  systems.
\newblock \emph{Automatica}, 36\penalty0 (4):\penalty0 595--601, 2000.

\bibitem[Landau \& Lifshitz(1976)Landau and Lifshitz]{landau1976mechanics}
Lev~D. Landau and Evgeny~M. Lifshitz.
\newblock \emph{Mechanics}, volume~1 of \emph{Course of Theoretical Physics}.
\newblock Butterworth-Heinemann, 3 edition, 1976.
\newblock \S 27, parametric resonance.

\bibitem[Lin \& Cai(2004)Lin and Cai]{lin2004}
Y.~K. Lin and G.~Q. Cai.
\newblock \emph{Probabilistic Structural Dynamics: Advanced Theory and
  Applications}.
\newblock McGraw-Hill, 2004.

\bibitem[Markus \& Yamabe(1960)Markus and Yamabe]{markus1960}
Lawrence Markus and Hidehiko Yamabe.
\newblock Global stability criteria for differential systems.
\newblock \emph{Osaka Mathematical Journal}, 12\penalty0 (2):\penalty0
  305--317, 1960.

\bibitem[Nayfeh \& Mook(1979)Nayfeh and Mook]{nayfeh1979}
Ali~H. Nayfeh and Dean~T. Mook.
\newblock \emph{Nonlinear Oscillations}.
\newblock Wiley, 1979.
\newblock \S 5.3.4, the {M}athieu equation.

\bibitem[Panda \& Bhatnagar(2024)Panda and Bhatnagar]{panda2023}
Prashansa Panda and Shalabh Bhatnagar.
\newblock Finite-time analysis of three-timescale constrained actor-critic and
  constrained natural actor-critic algorithms.
\newblock In \emph{Proceedings of the 27th International Conference on
  Artificial Intelligence and Statistics (AISTATS)}, volume 244 of
  \emph{Proceedings of Machine Learning Research}, 2024.
\newblock arXiv:2310.16363.

\bibitem[Paternain et~al.(2019)Paternain, Chamon, Calvo-Fullana, and
  Ribeiro]{paternain2019}
Santiago Paternain, Luiz Chamon, Miguel Calvo-Fullana, and Alejandro Ribeiro.
\newblock Constrained reinforcement learning has zero duality gap.
\newblock In \emph{Advances in Neural Information Processing Systems},
  volume~32, 2019.
\newblock arXiv:1910.13393.

\bibitem[Platt \& Barr(1988)Platt and Barr]{platt1988}
John~C. Platt and Alan~H. Barr.
\newblock Constrained differential optimization.
\newblock In \emph{Neural Information Processing Systems}, pp.\  612--621,
  1988.
\newblock Proceedings of NIPS 1987.

\bibitem[Rand(2012)]{rand2012}
Richard~H. Rand.
\newblock Lecture notes on nonlinear vibrations, 2012.
\newblock URL \url{https://ecommons.cornell.edu/handle/1813/28989}.
\newblock Cornell University, chapter on the {M}athieu equation.

\bibitem[Ray et~al.(2019)Ray, Achiam, and Amodei]{ray2019}
Alex Ray, Joshua Achiam, and Dario Amodei.
\newblock Benchmarking safe exploration in deep reinforcement learning.
\newblock Technical report, OpenAI, 2019.
\newblock Technical report accompanying the Safety Gym benchmark.

\bibitem[Rohrs et~al.(1985)Rohrs, Valavani, Athans, and Stein]{rohrs1985}
Charles~E. Rohrs, Lena Valavani, Michael Athans, and Gunter Stein.
\newblock Robustness of continuous-time adaptive control algorithms in the
  presence of unmodeled dynamics.
\newblock \emph{IEEE Transactions on Automatic Control}, 30\penalty0
  (9):\penalty0 881--889, 1985.

\bibitem[Simonetto et~al.(2020)Simonetto, Dall'Anese, Paternain, Leus, and
  Giannakis]{simonetto2020}
Andrea Simonetto, Emiliano Dall'Anese, Santiago Paternain, Geert Leus, and
  Georgios~B. Giannakis.
\newblock Time-varying convex optimization: Time-structured algorithms and
  applications.
\newblock \emph{Proceedings of the IEEE}, 108\penalty0 (11):\penalty0
  2032--2048, 2020.

\bibitem[Stooke et~al.(2020)Stooke, Achiam, and Abbeel]{stooke2020}
Adam Stooke, Joshua Achiam, and Pieter Abbeel.
\newblock Responsive safety in reinforcement learning by {PID} {L}agrangian
  methods.
\newblock In \emph{Proceedings of the 37th International Conference on Machine
  Learning}, volume 119 of \emph{Proceedings of Machine Learning Research},
  pp.\  9133--9143, 2020.

\bibitem[Stratonovich(1967)]{stratonovich1967}
Ruslan~L. Stratonovich.
\newblock \emph{Topics in the Theory of Random Noise, Volume II}.
\newblock Gordon and Breach, 1967.
\newblock Translated from Russian by Richard A. Silverman.

\bibitem[Tang et~al.(2022)Tang, Dall'Anese, Bernstein, and Low]{tang2022}
Yujie Tang, Emiliano Dall'Anese, Andrey Bernstein, and Steven~H. Low.
\newblock Running primal-dual gradient method for time-varying nonconvex
  problems.
\newblock \emph{SIAM Journal on Control and Optimization}, 60\penalty0
  (4):\penalty0 1970--1990, 2022.

\bibitem[Tessler et~al.(2019)Tessler, Mankowitz, and Mannor]{tessler2019}
Chen Tessler, Daniel~J. Mankowitz, and Shie Mannor.
\newblock Reward constrained policy optimization.
\newblock In \emph{International Conference on Learning Representations
  (ICLR)}, 2019.

\end{thebibliography}
\bibliographystyle{tmlr}

\appendix

\section{Extended setup and discussion}\label{sec:ec_extended}

This section collects the notation and the setup material that the body states in short form.

\subsection{Notation}\label{sec:notation}

Symbols are fixed once here and used throughout, in the convention of \citet{baelde2026companion}. Vectors and matrices are set in bold ($\boldsymbol\theta$, $\mathbf A_L$, $\mathbf c$) and appear only in Section~\ref{sec:setup_cmdp} and in Remark~\ref{rem:multiK}; every symbol below, in particular $a$ and $c$, is already the scalar reduction along the constraint direction (Section~\ref{sec:setup_cmdp}).

\begin{table}
\TABLE
{Table of notation\label{tab:notation}}
{\resizebox{\textwidth}{!}{%
\begin{tabular}{ll@{\quad}ll}
$\theta$ & policy parameter, tangent coordinate & $\omega$ & frequency of the non-stationarity \\
$\lambda$ & multiplier of the active constraint & $A$ & amplitude of the additive forcing \\
$\etath$ & primal step size & $\wnat$, $\wex$ & natural frequency, continuous and exact \\
$\etalam$ & dual step size & $\sigma$ & damping rate of the pair \\
$\rratio = \etalam/\etath$ & timescale ratio & $\zeta = \sigma/\wnat$ & reduced damping \\
$\ghat$ & constraint estimate, consumed by the primal & $\epsth$ & modulation depth seen by the primal \\
$g$ & constraint measurement, consumed by the dual & $\epslam$ & modulation depth seen by the dual \\
$a$ & curvature of the objective at the saddle & $\Delta\lambda$ & peak-to-peak amplitude of the multiplier \\
$c$ & constraint sensitivity $\partial g/\partial\theta$ & $B$ & bandwidth of the mismatch spectrum \\
$b$ & constant part of the constraint level & $K$ & number of active constraints \\
$h = \epslam - \epsth$ & mismatch between the two sensitivity paths & $s_t$ & normalized modulation signal, $|s_t| \le 1$ \\
$\varepsilon$ & generic (coherent) modulation depth & $\Delta\varphi$ & phase offset between the two modulation paths \\
$Z$ & complex envelope of the resonant component of $\theta$ & $\mu$ & Floquet exponent, or top Lyapunov exponent \\
$\epsilon_0$ & narrowband threshold, root-mean-square convention & $\kappa$ & broadband tongue-broadening constant \\
\end{tabular}%
}}
{}
\end{table}

The \emph{transfer function} $H_\lambda(z)$ of Section~\ref{sec:additive} is the ratio, in the $z$-domain, of the multiplier response to the constraint disturbance that drives it, and exists only because the additive channel leaves the linearized dynamics time-invariant.

\subsection{Slow variation and the canonical two-timescale setting}\label{sec:ec_setup_extra}

Assumption~\ref{ass:slow} is what makes \eqref{eq:functionals} a legitimate description of the environment at time $t$ with $\xi = \xi_t$: the occupancy measure has time to equilibrate between two appreciable changes of the exogenous process, so the learner faces a slowly moving constrained optimization problem. It is also the regime of the motivating application, since a daily traffic cycle is orders of magnitude slower than the horizon of an individual bidding episode.

The separation required by the two-timescale theorem of \citet[ch.~6]{borkar2008} is written $b(n)/a(n) \to 0$ in Borkar's own notation, with $a(n)$ the faster schedule and $b(n)$ the slower one. That notation is unrelated to our curvature $a$ and level constant $b$; $a(n)$ and $b(n)$ correspond here to $\etath$ and $\etalam$ respectively.

Three deviations from the canonical two-timescale setting determine what kind of statement is available. We work at constant step sizes, without $\etath, \etalam \to 0$, so convergence is weak, to a stationary law of dispersion $\mathcal O(\eta)$, and the available question is the sign of an exponent. Our ratio $\rratio$ is held fixed, without asymptotic separation, so a quasi-static approximation stands in for the theorem's timescale separation, its error measured by $\zeta$ itself. Third, and decisively, non-stationary $\xi_t$ leaves the fast ODE at frozen slow variable with no fixed equilibrium, so global asymptotic stability must be replaced by stability uniform in $t$, requiring a uniform Lyapunov function \citep[ch.~4 and 9]{khalil2002}.

The step sizes of Table~\ref{tab:deployed} are read from the repositories in August 2026: \texttt{pi\_lr}/\texttt{penalty\_lr} in \texttt{safe\_rl/pg/agents.py} and \texttt{run\_agent.py}; \texttt{actor.lr}/\texttt{lagrange\_cfgs.lambda\_lr} in \texttt{omnisafe/configs/on-policy/PPOLag.yaml}; the hardcoded actor rate and \texttt{--lagrangian-multiplier-lr} in \texttt{safepo/single\_agent/ppo\_lag.py} and \texttt{safepo/utils/config.py}. All three step the multiplier once per rollout over many policy gradient steps on the same batch, so the tabulated $\rratio$ is an upper bound; what survives is the sign of the comparison. The primal's constraint signal is a per-step cost advantage from a learned cost critic (\texttt{adv\_c} in \texttt{omnisafe}/\texttt{Safe-Policy-Optimization}, the penalized advantage in \texttt{safety-starter-agents}), against the dual's raw batch-averaged empirical cost (\texttt{EpCost}).

With $c \approx 1$ and the \texttt{omnisafe} \texttt{PPOLag} step sizes of Table~\ref{tab:deployed}, $\wnat \approx 3.2\times10^{-3}$ per update by \eqref{eq:wnat}, so the resonance condition $\omega = 2\wnat$ corresponds to a period of $N^\star = \pi/\wnat \approx 970$ updates, i.e.\ 16 hours to 1 week at a cadence of 1 to 10 minutes per update, the same order of magnitude as the diurnal cycle.

\subsection{The parametric channel: complements}\label{sec:ec_param_extra}

\paragraph{The dropped offset term.} In deviation coordinates around the unmodulated saddle $(\theta^\star,\lambda^\star) = (-b/c, ab/c^2)$, modulating a sensitivity also modulates its product with the offset, so each path contributes an additive term proportional to $b$: an input $-\etath ab\epsth s_t/c$ on the primal line and $-\etalam b\,\epslam s_t$ on the dual line. Both are exogenous inputs at the modulation frequency, that is the additive channel of Section~\ref{sec:additive}, bounded by Proposition~\ref{prop:floor} uniformly in their amplitude. They shift the trajectory but cannot change the sign of an exponent, which is decided by the homogeneous part \eqref{eq:stepmatrix} alone, and that part is what the Floquet and Lyapunov statements analyze. This holds while the projection is inactive. The additive response is bounded but not small, of amplitude $\etalam c|\theta^\star|\,\epslam\,\omega/|\omega^2 - \wex^2|$ on the multiplier when $\etath a \ll \omega$, and in the agent of Section~\ref{sec:exp_deeprl} it reaches $\lambda^\star$ near the tongue: the projection then becomes active, and the resulting saturation, which the homogeneous part does not see, decides whether the sub-harmonic locks.

\paragraph{General periodic waveform under coherence.} For a general $T$-periodic $s$ with $k\,\omega = 2\wnat$, the pump in Proposition~\ref{thm:coherent}(ii) is $\sigma|c_k|$, where $c_k$ is the $k$-th Fourier coefficient of $\exp\big(-2i\wnat\varepsilon\int_0^t s\big)$, and $|c_k| < 1$. The proof of (i) is a direct computation on the weighted energy; (ii) and this extension follow from an exact change of variables to the frame of the instantaneous phase, in which a coherent modulation becomes a pure phase modulation and the only forcing left acting on the conjugate envelope is the damping itself. The Bessel coefficient is the Jacobi--Anger expansion of that phase.

\paragraph{The discrete recursion under coherence.} The step matrix \eqref{eq:stepmatrix} inherits Proposition~\ref{thm:coherent}(ii) with two corrections of its own. Its determinant $\det M_t = 1 - \etath a + \wnat^2(1 + \varepsilon s_t)^2$ varies along the period, so the exponent carries the exact mean $\frac{1}{2N}\sum_t \log\det M_t$ in place of $-\sigma$, and the pump coefficient is $\etath a/2$ in place of $\sigma$; the exponent is then $\frac{1}{2N}\sum_t\log\det M_t + \operatorname{Re}\sqrt{(\etath a/2)^2 J_1(\beta)^2 - \Delta^2}$ with $\Delta = \omega/2 - \wex$. The condition of Remark~\ref{rem:secondorder} fails as the pair $(\etath a, \etalam c^2)$ approaches the boundary of the Jury region \eqref{eq:jury}, where the instantaneous step matrix at the peak of the modulation leaves its own stability region. Every operating point in this paper satisfies it with a wide margin, and the measured exponents of Section~\ref{sec:exp_narrowband} follow the formula to the second order in $\zeta$. The exact mean and the pump coefficient are derived in Appendix~\ref{sec:ec_proof_coherent}.

\paragraph{On the threshold.} The naive closed form $4\zeta \simeq (2a/c)\sqrt{\etath/\etalam}$ obtained by dropping the correction term of \eqref{eq:threshold_closed} is a first-order approximation, and the tolerance collapses as $\sqrt{\etath/\etalam}$. The averaging is carried out on the continuous surrogate, of natural frequency $\wnat$, while the tongue of the recursion \eqref{eq:stepmatrix} sits at twice the argument $\wex$ of \eqref{eq:sigma_omega}. The two differ at relative order $\zeta^2$, below the tongue's half-width and above measurement precision, so the statement is written with $\wnat$ and the experiments sweep the modulation period around $\pi/\wex$. The statement is narrowband, requiring the mismatch spectrally concentrated on a band of width $B \ll \sigma/\kappa$ around the tongue ($\kappa$ the constant of Proposition~\ref{prop:spectral}); the broadband case returns \eqref{eq:threshold} at $B \to 0$. The proof decomposes the envelope of Lemma~\ref{lem:cancellation} into the two decoupled scalar equations that give $\mu = \tfrac{\wnat}{4}|h| - \sigma$, then carries the same argument through the discrete monodromy of \eqref{eq:stepmatrix}.

\paragraph{On the existence condition.} The two halves of \eqref{eq:existence} differ in status. Below $(a/c)^2/4$ it is exact and structural by \eqref{eq:underdamped} (no oscillatory mode, no mechanism). Above it, up to $(a/c)^2$, it rests on the first-order threshold \eqref{eq:threshold} alone, evaluated at the physical maximum $|h|=2$ rather than at the small depth the expansion is derived for. It is mildly optimistic on that count, and also because the tongue tilts away from the unmodulated centre frequency as the mismatch grows (Appendix~\ref{sec:ec_controls} quantifies the tilt). The proof bounds $|h| \le 2$ from the physical depth constraint $|\varepsilon| \le 1$ and crosses it against the first-order threshold $4\zeta = (2a/c)\rratio^{-1/2}$.

\paragraph{The broadband averaging.} The averaged growth rate behind Proposition~\ref{prop:spectral} is $\mu = -\sigma + \frac{\wnat^2}{32}S_m(0)$, where $S_m$ is the power spectral density of the mismatch process at zero offset from the tongue centre, two-sided, in angular frequency, and taken of the complex envelope of the mismatch, normalised so that $\frac{1}{2\pi}\int S_m(\omega)\,d\omega = 2\varepsilon^2$. For a mismatch of total mean-square depth $\varepsilon^2$ spread uniformly over a band of width $B$, $S_m(0) = 4\pi\varepsilon^2/B$, which substituted gives \eqref{eq:spectral_asymptote} and, equivalently, \eqref{eq:broadband}. The crossover shape the interpolation \eqref{eq:spectral} assumes at intermediate $B$ is not itself derived from the averaging argument.

\begin{remark}[Active constraints and gradient noise]\label{rem:multiK}
Let $K$ constraints be active, with constraint Jacobian $C \in \mathbb{R}^{K\times d}$ of full row rank whose row space is invariant under the Lagrangian curvature and isotropic ($\mathbf A_L = aI$) on it and its complement, the multidimensional condition under which the single number $a$ is exact. Under a single waveform scaling every sensitivity ($C$ by $1+\epsth s_t$ primal, $1+\epslam s_t$ dual), the linearization decouples \emph{exactly} in the singular basis of $C$. It splits into $K$ copies of the scalar system above, the $k$-th carrying $c^2 = \varsigma_k^2$ for $\varsigma_k$ the $k$-th singular value of $C$. Tongues sit at $2\varsigma_k\sqrt{\etath\etalam}$, and the binding tolerance $4\zeta = (2a/\varsigma_{\max})\sqrt{\etath/\etalam}$ is set by the largest singular value alone. It is unchanged by orthogonal constraints of equal norm, and falls as $K^{-1/2}$ for nearly collinear ones. Heterogeneous \emph{frequencies} fall outside the reduction: several waveforms address several tongues, and nothing above says how they interact.

Gradient noise reduces as easily. Additive zero-mean noise moves the stationary variance without touching the Floquet or Lyapunov exponents, since the threshold is a property of the homogeneous part \eqref{eq:stepmatrix} alone. State-dependent noise, which enters \eqref{eq:stepmatrix} itself, falls outside the tangent reduction.
\end{remark}

\section{Deferred proofs}\label{sec:ec_proofs}

This section holds the proofs that the body defers: the low-frequency floor and worst-case amplification of the additive channel, the averaging computation behind the damping-stiffness cancellation, the narrowband mismatch threshold that the cancellation feeds into, the coherent-modulation theorem, the broadband scaling law, and the existence corollary. Equation and statement numbers below refer to the body.

\subsection{The additive channel: floor and worst-case amplification}\label{sec:ec_proof_additive}

Besides \eqref{eq:jury}, the Jury criterion carries the two side conditions $\etath(2a - \etalam c^2) < 4$ and $\etath(a - \etalam c^2) < 2$. These are the requirements that a single step not overshoot, and they hold with a wide margin at every operating point in this paper. The threshold $(a/c)^2$ of \eqref{eq:underdamped} is itself unitful, so every comparison of $\rratio$ against it, including the orders of magnitude read off Table~\ref{tab:deployed}, implicitly holds $a/c$ at the $\mathcal O(1)$ scale of the experiments of Section~\ref{sec:experiments}.

In Proposition~\ref{prop:floor}(i) the plateau extends up to $\omega_c = \etalam c^2/a = \wnat/(2\zeta)$ when $\zeta > 1/2$ and up to the resonance $\wnat$ when $\zeta < 1/2$. The threshold $\zeta = 1/2$ separating the two branches is the one at which the dual corner $\etalam c^2/a$ meets $\wnat$. It is distinct from the threshold $\zeta = 1$ at which the poles become complex, which is \eqref{eq:underdamped} rewritten through \eqref{eq:wnat}. In (ii), the second reduction carries a relative error $\etalam c^2/a$ that does not vanish with $\zeta$, so the first form is the one to compare a measurement against. On the intermediate range $1/2 < \zeta < 1$ the poles are already complex but the resonance does not rise above the floor of (i); the supremum is still attained as $\omega \to 0$, so the maximum is not redundant.

\begin{proof}{Proof of Proposition~\ref{prop:floor}(i)}
Evaluate \eqref{eq:transfer} at $z = 1$: the numerator is $\etath a$ and the denominator is $\etath\etalam c^2$, so $H_\lambda(1) = a/(\etalam c^2)$ exactly, and $\Delta\lambda \to 2\etalam A\cdot a/(\etalam c^2) = 2Aa/c^2$, in which $\etalam$ cancels. For the extent of the plateau, note that for $\omega \ll \etath a$ the numerator of \eqref{eq:transfer} is $\etath a(1 + \mathcal O(\omega/\etath a))$ and the denominator is $\etath a\,i\omega + \etath\etalam c^2 + \mathcal O(\omega^2)$, so $H_\lambda(e^{i\omega}) \simeq a/(a\,i\omega + \etalam c^2)$, a single pole with corner $\omega_c = \etalam c^2/a$. That approximation is valid only while $\omega_c \ll \etath a$, which by \eqref{eq:wnat} is $\zeta \gg 1/2$; in the complementary underdamped case the ratio $\omega_c/\wnat = (c/a)\sqrt{\etalam/\etath} = (2\zeta)^{-1}$ exceeds one, the second pole is met first, and the plateau ends at $\wnat$ where the resonant peak sits. \Halmos
\end{proof}

\begin{proof}{Proof of Proposition~\ref{prop:worstcase}(ii)}
Evaluate \eqref{eq:transfer} on the unit circle near the resonance, $z = e^{i\omega}$ with $\omega = \wnat + \mathcal O(\zeta\wnat)$, and write $u = e^{i\omega} - 1 = i\omega - \omega^2/2 + \mathcal O(\omega^3)$. The numerator is $u + \etath a$, of modulus $\wnat\sqrt{1 + 4\zeta^2} = \wnat(1 + \mathcal O(\zeta^2))$ by \eqref{eq:wnat}. The denominator is $u^2 + \etath a\,u + \wnat^2$, and since $u^2 = -\omega^2 - i\omega^3 + \mathcal O(\omega^4)$ the term $\wnat^2 = \etath\etalam c^2$ cancels the leading real part and leaves
\[
i\omega\big(\etath a - \etath\etalam c^2\big) - \tfrac12\etath a\,\omega^2 + \mathcal O(\omega^4) = 2\sigma\,i\omega\,\big(1 + \mathcal O(\etath)\big),
\]
the discarded real part being smaller than the retained imaginary one by a factor $\mathcal O(\wnat)$. Hence $|H_\lambda|_{\max} \simeq 1/(2\sigma)$ and $\sup_\omega \Delta\lambda \simeq A\etalam/\sigma$. The cubic term in $u^2$ is what turns $\etath a$ into $2\sigma$: dropping it replaces the damping of the pole by the primal contribution to it alone, an error of relative size $\etalam c^2/a$. Dividing by the floor $2Aa/c^2$ of Proposition~\ref{prop:floor} gives $\rratio c^2/a^2 = (2\zeta)^{-2}$, which exceeds one exactly when $\rratio > (a/c)^2/4$, that is under \eqref{eq:underdamped}; in the overdamped case the response is monotone and the supremum is the floor. \Halmos
\end{proof}

\subsection{The damping-stiffness cancellation}\label{sec:ec_proof_lemma}

\begin{proof}{Proof of Lemma~\ref{lem:cancellation}}
Start from \eqref{eq:mathieu_raw} with $s_t = \sin\omega t$, so that $\dot c_\theta/c_\theta = \epsth\omega\cos\omega t + \mathcal O(\epsth^2)$ and $\etath\etalam c_\theta c_\lambda = \wnat^2\big(1 + (\epsth + \epslam)\sin\omega t\big) + \mathcal O(\varepsilon^2)$. Writing $2\sigma = \etath a$ to leading order in the step sizes, \eqref{eq:mathieu_raw} becomes
\begin{equation}\label{eq:ec_mathieu_first}
\ddot\theta + 2\sigma\dot\theta + \wnat^2\theta = -\wnat^2(\epsth + \epslam)\sin(\omega t)\,\theta + \epsth\omega\cos(\omega t)\,\dot\theta + 2\sigma\epsth\omega\cos(\omega t)\,\theta + \mathcal O(\varepsilon^2).
\end{equation}
The last term on the right is $\mathcal O(\zeta\varepsilon)$ relative to the first two, and at the joint order the lemma is stated to, first order in $(\epsth,\epslam)$ \emph{and} in $\zeta$ together, the modulation depth is itself $\mathcal O(\zeta)$, so that term is of the same order as the neglected $\mathcal O(\varepsilon^2)$ and is dropped here. That order pairing is exactly the regime Theorem~\ref{thm:threshold} uses the lemma in, and nothing below revisits it. Two forcing terms remain at this order: a stiffness modulation of depth $\epsth + \epslam$, contributed by both paths through the product $c_\theta c_\lambda$, and a damping modulation of depth $\epsth$, contributed by the primal path alone through $\dot c_\theta/c_\theta$. Appendix~\ref{sec:ec_proof_coherent} below treats the coherent case at a depth $\varepsilon$ of order one, where $\varepsilon$ and $\zeta$ are no longer tied together and the pairing that licenses dropping that term no longer holds; it leaves the frame at $\omega/2$ for the frame of the instantaneous phase, in which no term has to be dropped at any depth.

Substitute $\theta(t) = \tfrac12\big(Ze^{i\omega t/2} + \bar Ze^{-i\omega t/2}\big)$ with $Z$ slowly varying, so that $\dot\theta = \tfrac12\big(i\tfrac\omega2 Z + \dot Z\big)e^{i\omega t/2} + \mathrm{c.c.}$ and $\ddot\theta = \tfrac12\big(-\tfrac{\omega^2}{4}Z + i\omega\dot Z\big)e^{i\omega t/2} + \mathrm{c.c.}$, discarding $\ddot Z$. Collecting the coefficient of $e^{i\omega t/2}$ on the left of \eqref{eq:ec_mathieu_first} gives
\[
\tfrac12\Big[i\omega\dot Z + \big(\wnat^2 - \tfrac{\omega^2}{4}\big)Z + i\sigma\omega Z\Big],
\]
which at $\omega = 2\wnat$ is $i\wnat(\dot Z + \sigma Z)$.

On the right, each forcing term is a product of a modulation at $\omega$ with a signal at $\omega/2$, and only the difference frequency contributes to the resonant component. For the stiffness term, $\sin\omega t = \tfrac{1}{2i}(e^{i\omega t} - e^{-i\omega t})$ beats against the $\bar Ze^{-i\omega t/2}$ half of $\theta$ and yields the coefficient $-\tfrac{i}{4}\bar Z$ for $e^{i\omega t/2}$, so the term $-\wnat^2(\epsth+\epslam)\sin(\omega t)\theta$ contributes $\tfrac{i}{4}\wnat^2(\epsth + \epslam)\bar Z$. For the damping term, $\cos\omega t = \tfrac12(e^{i\omega t} + e^{-i\omega t})$ beats against the $\overline{(i\tfrac\omega2 Z)}e^{-i\omega t/2}$ half of $\dot\theta$ and yields the coefficient $-\tfrac{i}{8}\omega\bar Z$, so the term $\epsth\omega\cos(\omega t)\dot\theta$ contributes $-\tfrac{i}{8}\epsth\omega^2\bar Z$. Summing,
\[
i\wnat\big(\dot Z + \sigma Z\big) = \frac{i}{4}\Big[\wnat^2(\epsth + \epslam) - \frac{\omega^2}{2}\epsth\Big]\bar Z,
\]
which is the envelope equation of the lemma after division by $i\wnat$, and the bracket is \eqref{eq:bracket}. At $\omega = 2\wnat$ the bracket is $\wnat^2\big[(\epsth+\epslam) - 2\epsth\big] = \wnat^2(\epslam - \epsth)$, so the two contributions have equal magnitude and opposite sign there. \Halmos
\end{proof}

\subsection{The narrowband mismatch threshold}\label{sec:ec_proof_threshold}

\begin{proof}{Proof of Theorem~\ref{thm:threshold}}
Write $h = \epslam - \epsth$ and decompose the envelope of Lemma~\ref{lem:cancellation} into real and imaginary parts, $Z = Z_1 + iZ_2$, at $\omega = 2\wnat$. Since the forcing acts on $\bar Z$, the envelope equation $\dot Z = -\sigma Z + \tfrac{\wnat}{4}h\bar Z$ splits into two decoupled scalar equations, $\dot Z_1 = (-\sigma + \tfrac{\wnat}{4}h)Z_1$ and $\dot Z_2 = (-\sigma - \tfrac{\wnat}{4}h)Z_2$. The growth rate of the unstable direction is therefore $\mu = \tfrac{\wnat}{4}|h| - \sigma$, and $\mu > 0$ if and only if $|h| > 4\sigma/\wnat = 4\zeta$, which is \eqref{eq:threshold}. Substituting the exact $\sigma = -\tfrac12\log(1 - \etath a + \etath\etalam c^2) = \tfrac{\etath}{2}(a - \etalam c^2) + \mathcal O(\etath^2)$ of \eqref{eq:sigma_omega} for $\sigma$, in place of the leading term $\sigma \simeq \tfrac12\etath a$ of \eqref{eq:wnat} alone, gives the closed form \eqref{eq:threshold_closed}; keeping only the leading term recovers the naive $4\zeta \simeq (2a/c)\sqrt{\etath/\etalam}$, which is accurate only to $\mathcal O(\etalam c^2/a)$.

Replacing $\wnat$ by $\wex$ perturbs $\mu$ at relative order $\zeta^2$, below the order at which the expansion holds, so the choice between the two frequencies is immaterial to \eqref{eq:threshold} and material only to the period at which the exponent is measured, which is the one of the recursion.

For the discrete system the conclusion follows from the monodromy $\Phi = \prod_{t<N}M_t$ of \eqref{eq:stepmatrix} over an integer period $N$, $\omega = 2\pi/N$, by the argument of Proposition~\ref{prop:ec_coherent_discrete} below carried through at a mismatched depth. In the coordinates $x_t = \sqrt{\etalam}\theta_t$, $y_t = \sqrt{\etath}\lambda_t$ of that proposition's proof, the recursion reads $x_{t+1} = (1-\etath a)x_t - \Omega_{\theta,t}y_t$, $y_{t+1} = y_t + \Omega_{\lambda,t}x_t$ with $\Omega_{\theta,t} = \wnat(1+\epsth s_t)$ and $\Omega_{\lambda,t} = \wnat(1+\epslam s_t)$, since $c\sqrt{\etath\etalam} = \wnat$. Split the two gains into their common and differential parts,
\begin{equation}\label{eq:ec_zsplit}
\Omega_{\theta,t} = \Omega_t - \delta_t, \qquad \Omega_{\lambda,t} = \Omega_t + \delta_t, \qquad
\Omega_t = \wnat(1 + \bar\varepsilon s_t), \quad \bar\varepsilon = \tfrac12(\epsth+\epslam), \quad \delta_t = \tfrac{\wnat h}{2}s_t,
\end{equation}
with $h = \epslam - \epsth$. Then $z_t = x_t + iy_t$ obeys, exactly and with no expansion in the depths,
\begin{equation}\label{eq:ec_zrec}
z_{t+1} = (1 + i\Omega_t)z_t - \tfrac{\etath a}{2}(z_t + \bar z_t) + i\delta_t\bar z_t ,
\end{equation}
by $-\Omega_t y_t + i\Omega_t x_t = i\Omega_t z_t$, $\ \delta_t y_t + i\delta_t x_t = i\delta_t \bar z_t$ and $x_t = \tfrac12(z_t+\bar z_t)$. So the instantaneous frequency $\Omega_t$ carries the common depth $\bar\varepsilon$ alone, and the mismatch enters through a single additional conjugate term of coefficient $\wnat h/2$, of the same order as the damping's own conjugate term $\etath a/2 = \zeta\wnat + \mathcal O(\etath^2)$ exactly when $|h|$ is of order $\zeta$. Setting $h = 0$ returns the recursion of Proposition~\ref{prop:ec_coherent_discrete}.

The mean of the two exponents is exact and comes from the determinant: $\det M_t = 1 - \etath a + \wnat^2(1+\epsth s_t)(1+\epslam s_t)$, and $1 - \etath a + \wnat^2 = e^{-2\sigma}$ by \eqref{eq:sigma_omega}, so with $\langle s_t\rangle = 0$ and $\langle s_t^2\rangle = \tfrac12$ over the period,
\[
\frac{1}{2N}\sum_{t<N}\log\det M_t = -\sigma + \frac{\wnat^2}{4}\epsth\epslam + \mathcal O(\etath^2),
\]
which depends on the two depths only through their product, is $\mathcal O(\wnat)$ relative to $\sigma$, and returns $-\sigma$ of \eqref{eq:sigma_omega} at $\varepsilon = 0$. Under a pure mismatch, $\epsth\epslam \le 0$, it is a stabilizing correction.

The splitting of the pair about that mean is governed by the two conjugate terms of \eqref{eq:ec_zrec}. Pass to the frame of the discrete phase, $z_t = R_te^{i\phi_t}w_t$ with $\phi_t = \sum_{u<t}\arctan\Omega_u$ and $R_t = \prod_{u<t}\sqrt{1+\Omega_u^2}$, which turns \eqref{eq:ec_zrec} into $w_{t+1} - w_t = -\tfrac{\etath a}{2}w_t + \big(i\delta_t - \tfrac{\etath a}{2}\big)e^{-2i\phi_t}\bar w_t$; dividing through by $\sqrt{1+\Omega_t^2}\,e^{i\arctan\Omega_t} = 1 + \mathcal O(\wnat)$ perturbs a right-hand side already $\mathcal O(\zeta\wnat)$ and is dropped, as is the $\mathcal O(\wnat)$ discrepancy between $\sum_{u<t}s_u$ and $\int_0^t s$. For $s_t = \sin\omega t$ this gives $2\phi_t = 2\wex t + \beta(1-\cos\omega t) + \mathcal O(\wnat)$ with $\beta = 2\bar\varepsilon\wnat/\omega$, since $\arctan\wnat = \wex$ to relative order $\zeta^2$. Write $\omega = 2(\wex+\Delta)$, $w_t = Z_te^{i\Delta t}$ and expand by Jacobi--Anger, $e^{i\beta\cos\omega t} = \sum_n i^nJ_n(\beta)e^{in\omega t}$. Three terms of $\big(i\delta_t - \tfrac{\etath a}{2}\big)e^{-2i\phi_t - 2i\Delta t}$ are free of a fast phase, in place of the single one of the coherent case: $n = 1$ against the damping, and $n = 0$ and $n = 2$ against the two sidebands of $\delta_t$. Averaging over the period,
\begin{align}
Z_{t+1} - Z_t &= -\Big(i\Delta + \tfrac{\etath a}{2}\Big)Z_t + F\bar Z_t, \notag\\
F &= e^{-i\beta}\Big[-\tfrac{i\etath a}{2}J_1(\beta) + \tfrac{\wnat h}{4}\big(J_0(\beta)+J_2(\beta)\big)\Big]
= e^{-i\beta}J_1(\beta)\Big[\frac{\wnat h}{2\beta} - \frac{i\etath a}{2}\Big], \label{eq:ec_disc_forcing}
\end{align}
by $J_0(\beta) + J_2(\beta) = 2J_1(\beta)/\beta$. The coherent and mismatched contributions arrive in quadrature, which is the discrete form of the orthogonality Lemma~\ref{lem:cancellation} exhibits as a cancellation of signs at the tongue centre. Splitting $Z$ into real and imaginary parts gives a constant real rate matrix, whence, with the exact mean above in place of its leading term $-\sigma$,
\begin{equation}\label{eq:ec_disc_mu}
\mu = \frac{1}{2N}\sum_{t<N}\log\det M_t + \operatorname{Re}\sqrt{J_1(\beta)^2\Big[\Big(\frac{\wnat h}{2\beta}\Big)^2 + \Big(\frac{\etath a}{2}\Big)^2\Big] - \Delta^2},
\qquad \frac{J_1(\beta)}{\beta}\Big|_{\beta = 0} = \frac12 ,
\end{equation}
which reduces to Proposition~\ref{prop:ec_coherent_discrete} at $h = 0$ and is symmetric in $\epsth \leftrightarrow \epslam$. Note that the conjugate coefficient carries $\etath a/2$ while the mean carries the exact $\sigma$ of \eqref{eq:sigma_omega}, the two differing at relative order $\etalam c^2/a$; the distinction is the same one Proposition~\ref{prop:worstcase} turns on and it is not optional at the operating point of Table~\ref{tab:opoint}.

At the order of Theorem~\ref{thm:threshold}, $\epsth$, $\epslam$ and $\zeta$ of one common order, $\beta = \mathcal O(\zeta)$ so $J_1(\beta)/\beta = \tfrac12 + \mathcal O(\zeta^2)$, the coherent part $J_1(\beta)\etath a/2$ of $|F|$ is $\mathcal O(\zeta^2\wnat)$, and the determinant correction $\wnat^2\epsth\epslam/4$ is $\mathcal O(\wnat\zeta)$ relative to $\sigma$. At the tongue centre, $\Delta = 0$, \eqref{eq:ec_disc_mu} therefore collapses to $\mu = \tfrac{\wnat}{4}|h| - \sigma$, with $\wex$ in place of $\wnat$ at relative order $\zeta^2$ as above, and to \eqref{eq:threshold}. Beyond that order \eqref{eq:ec_disc_mu} retains two effects the continuous statement does not: a coherent residue of relative size $\zeta$, and a depression of the growth rate at fixed $h$ as the common depth $\bar\varepsilon$ grows, through $J_1(\beta)/\beta$ and through $\epsth\epslam$. Both are visible in Table~\ref{tab:ec_recursion}, whose six exponents \eqref{eq:ec_disc_mu} reproduces to $3.3\times10^{-7}$ absolute, $0.15\%$ of $|\mu|$ and of the order $\zeta^2\sigma$ of the discarded terms, where the leading form $\tfrac{\wnat}{4}|h|-\sigma$ leaves the three coherent rows flat at $-\sigma$ and misses the $1.2\%$ separation between the two mismatched rows of equal $h$.
\Halmos
\end{proof}

\subsection{Coherent modulation}\label{sec:ec_proof_coherent}

\begin{proof}{Proof of Proposition~\ref{thm:coherent}}
Write $x = \sqrt{\etalam}\,\theta$, $y = \sqrt{\etath}\,\lambda$, $\tilde\omega(t) = c_\theta(t)\sqrt{\etath\etalam} = \wnat(1 + \varepsilon s(t))$ and $2\sigma = \etath a$. Under coherence the flow reads $\dot x = -2\sigma x - \tilde\omega(t)y$, $\dot y = \tilde\omega(t)x$: the modulation multiplies the antisymmetric part of the generator only, and the damping is the only symmetric part.

(i) With $V = x^2 + y^2$, the cross terms cancel and $\dot V = -4\sigma x^2 \le 0$. If $s$ is $T$-periodic and $\|\Phi v\|_V = \|v\|_V$ for some $v \ne 0$, then $x \equiv 0$ on $[0, T]$, hence $\dot x \equiv 0$, hence $\tilde\omega y \equiv 0$, hence $y \equiv 0$ because $\tilde\omega > 0$ when $|\varepsilon| < 1$ and $|s| \le 1$. So $\|\Phi\|_V < 1$ and $\rho(\Phi) < 1$. Under a mismatch $c_\theta \ne c_\lambda$ the same computation gives $\dot V = -4\sigma x^2 + 2\etath\etalam(c_\lambda - c_\theta)\theta\lambda$, whose second term has no sign, and that term is the parametric channel.

(ii) Let $z = x + iy$, so that $\dot z = i\tilde\omega(t)z - \sigma(z + \bar z)$, and pass to the frame of the exact phase, $z = w\,e^{i\phi(t)}$ with $\dot\phi = \tilde\omega$. Then, exactly,
\[
\dot w = -\sigma w - \sigma e^{-2i\phi(t)}\bar w .
\]
For $s = \sin\omega t$, $2\phi(t) = 2\wnat t + \beta(1 - \cos\omega t)$ with $\beta = 2\wnat\varepsilon/\omega$. Write $\omega = 2(\wnat + \Delta)$ and $w = Z e^{i\Delta t}$; by the Jacobi--Anger expansion $e^{i\beta\cos\omega t} = \sum_n i^n J_n(\beta)e^{in\omega t}$, and the only term of $e^{-2i\phi}\bar w\,e^{-i\Delta t}$ without a fast phase is the $n = 1$ term, so that after averaging over a period
\[
\dot Z = -(\sigma + i\Delta)Z + F\bar Z, \qquad F = -i\sigma e^{-i\beta}J_1(\beta).
\]
Splitting $Z$ into real and imaginary parts gives a constant real rate matrix with eigenvalues $-\sigma \pm \sqrt{|F|^2 - \Delta^2}$, which is the claim. Averaging is legitimate because $\sigma$ and $\Delta$ are $\mathcal O(\zeta\wnat)$ while the discarded terms oscillate at multiples of $\omega$, so the error is $\mathcal O(\zeta^2\wnat)$; the depth $\varepsilon$ enters through $J_1(\beta)$ only and is not expanded. When $|\Delta| > \sigma J_1(\beta)$ the eigenvalues are a complex pair of common real part $-\sigma$, consistent with $\det\Phi = e^{-2\sigma T}$.

(iii) For a general periodic $s$, $2\phi(t) = 2\wnat t + 2\wnat\varepsilon\int_0^t s$, and the averaged forcing coefficient is $\sigma$ times the Fourier coefficient of $e^{-2i\wnat\varepsilon\int_0^t s}$ at the frequency $2\wnat = k\omega$; its modulus is at most $1$, with equality impossible for a non-constant phase.

The frame of the exact phase also locates the error one commits by working in the frame at $\omega/2$ of Lemma~\ref{lem:cancellation}: there the coherent modulation appears as a phase modulation $i\varepsilon\omega\sin(\omega t)Z$ of the envelope, together with the term $\tfrac{\varepsilon\omega}{4}\dot{\bar Z}$ of $\epsth\omega\cos(\omega t)\dot\theta$, which is of the same order $\sigma\varepsilon\wnat$ as the term $2\sigma\epsth\omega\cos(\omega t)\theta$ and halves it; the two together give the first term $\varepsilon/2$ of $J_1(\varepsilon)$. \Halmos
\end{proof}

\begin{proposition}[Coherent modulation of the recursion]\label{prop:ec_coherent_discrete}
Let $M_t$ be the step matrix \eqref{eq:stepmatrix} with $\epsth = \epslam = \varepsilon$, $s_t = \sin(2\pi t/N)$, and let $\Delta = \pi/N - \wex$, $\beta = 2\varepsilon\wex N/(2\pi)$. To first order in $\zeta$, the top Floquet exponent of $\Phi = \prod_{t<N} M_t$ is
\[
\mu = \frac{1}{2N}\sum_{t=0}^{N-1}\log\det M_t + \operatorname{Re}\sqrt{\Big(\frac{\etath a}{2}\Big)^2 J_1(\beta)^2 - \Delta^2},
\qquad \det M_t = 1 - \etath a + \wnat^2(1 + \varepsilon s_t)^2 .
\]
The first term equals $-\sigma + \wnat^2\varepsilon^2/4 + \mathcal O(\etath^2)$ and does not depend on the forcing frequency. When the period is an integer, $\Delta \ne 0$ in general, and the multipliers remain a complex pair with exponent equal to the first term alone for all $\varepsilon$ below the depth at which $\tfrac{\etath a}{2}J_1(\beta) = |\Delta|$.
\end{proposition}

\begin{proof}{Proof}
In the coordinates of the proof of Proposition~\ref{thm:coherent}, $z_{t+1} = (1 + i\Omega_t)z_t - \tfrac{\etath a}{2}(z_t + \bar z_t)$ with $\Omega_t = \wnat(1 + \varepsilon s_t)$. The mean of $\tfrac12\log\det M_t$ is the mean of the two exponents, and it is exact. The splitting of the pair around that mean is governed by the $\bar z$ term, whose coefficient is $\etath a/2$; its resonant component is obtained as in the continuous case, with the discrete phase $\phi_t = \sum_{u<t}\arctan\Omega_u$ replacing $\int\tilde\omega$, which to first order in $\zeta$ reproduces the continuous result with $\wex$ in place of $\wnat$. \Halmos
\end{proof}

Off resonance a coherent modulation acquires only a bounded pump, over a band narrower than the parametric tongue by a relative factor of order $\zeta$ in both bandwidth and pump strength, so the coherent residue is smaller than the mismatch tongue's in both dimensions and part (i) of the proposition shows that it never crosses zero for the continuous flow.

\subsection{Lyapunov exponent of a random matrix product, and the scaling argument}\label{sec:ec_proof_lyapunov}

\begin{proof}{Proof of Proposition~\ref{prop:spectral}}
When $s_t$ is a stationary process, in place of a periodic signal, the product of the matrices \eqref{eq:stepmatrix} is no longer periodic and monodromy is unavailable. The stability object is then the top Lyapunov exponent
\[
\mu = \lim_{T\to\infty}\frac{1}{T}\log\Big\|\prod_{t=0}^{T-1}M_t\Big\|,
\]
which exists almost surely and is deterministic by the Furstenberg-Kesten theorem, since the $M_t$ are stationary, ergodic and uniformly bounded with $\det M_t$ bounded away from zero. The numerical instrument of Section~\ref{sec:protocol} generates $s_t$ as a finite, $T$-periodic band-limited signal, in place of a draw from a stationary process run to $T\to\infty$: what it measures on one draw is the per-draw exponent, which converges to the almost-sure constant $\mu$ above only as $T\to\infty$, and the draw-to-draw dispersion quantified there (Appendix~\ref{sec:ec_controls}) is exactly the gap between the two at the finite $T$ used.

The narrowband endpoint is the threshold of Theorem~\ref{thm:threshold} in root-mean-square convention. At $B = 0$ the mismatch is a single line at $2\wex$ and the process is periodic, so the exponent is the Floquet exponent \eqref{eq:floquet_rate} and it changes sign at the peak depth $4\zeta$; the conversion of a peak depth into a root-mean-square one contributes the factor $\sqrt2$, whence $\epscrit(0) = \varepsilon_0 = 2\sqrt2\,\zeta$. That value is inherited from \eqref{eq:threshold} and is therefore exact to the same first order in $\zeta$, no better.

For the broadband endpoint, write the envelope equation of Lemma~\ref{lem:cancellation} at the tongue centre $\omega = 2\wnat$ with a broadband stationary mismatch $m(t)$ in place of the single-line forcing, $\dot Z = -\sigma Z + \frac{\wnat}{4}m(t)\bar Z$, where $m$ is zero-mean, stationary, ergodic, and spectrally supported on a band of width $B$ centred on $2\wex$ so that it is slow compared with $\sigma^{-1}$ only in the narrowband limit and fast compared with it once $B \gg \sigma$. This is the regime of Stratonovich-Khasminskii second-order stochastic averaging \citep{stratonovich1967,khasminskii1966}, developed for exactly this class of randomly perturbed linear oscillators and applied to the Lyapunov exponent of the random harmonic oscillator by \citet{arnold1986} and, in the structural-dynamics setting from which the present envelope equation borrows its normal form, by \citet{lin2004}. The averaging theorem gives, for the top Lyapunov exponent of $Z$, the closed form
\[
\mu = -\sigma + \frac{\wnat^2}{32}\,S_m(0),
\]
where $S_m(0)$ is the two-sided power spectral density of $m$ evaluated at zero offset from the tongue centre, i.e.\ at the frequency where the stochastic forcing beats resonantly against the natural frequency of the envelope; frequencies away from that offset average out of the first-order correction to $\mu$, which is the content of the averaging theorem in this setting.

For a mismatch of total mean-square depth $\varepsilon^2$ spread uniformly over a real, flat band of width $B$ centred on $2\wex$, the corresponding complex baseband spectral density (the convention under which $S_m(0)$ enters the envelope equation above) has height $S_m(0) = 4\pi\varepsilon^2/B$: the complex envelope carries twice the real per-channel variance, so its mean square is $2\varepsilon^2$ and its baseband support has width $B$; in the convention of $\mu$ above, two-sided and in angular frequency with $\frac{1}{2\pi}\int S_m(\omega)\,d\omega = 2\varepsilon^2$, a flat density of that mass over that width has height $S_m(0) = 4\pi\varepsilon^2/B$. Substituting,
\[
\mu = -\sigma + \frac{\wnat^2}{32}\cdot\frac{4\pi\varepsilon^2}{B} = -\sigma + \frac{\pi\wnat^2}{8B}\,\varepsilon^2,
\]
and $\mu$ changes sign at
\[
\varepsilon_{\mathrm{crit}}^2(B) = \frac{8\sigma B}{\pi\wnat^2}.
\]
Using $\varepsilon_0^2 = 8\zeta^2 = 8\sigma^2/\wnat^2$ from the narrowband endpoint above, this is $\epscrit^2(B) = \varepsilon_0^2\,B/(\pi\sigma)$, i.e.\ \eqref{eq:spectral_asymptote} with $\kappa = 1/\pi$; the value is set entirely by the coefficient $\wnat^2/32$ of the averaged growth rate, itself fixed by the profile of the envelope equation at the tongue, and does not involve either step size.

The cancellation of $\etath$ in \eqref{eq:broadband} follows by substitution. In the regime $B \gg \sigma$, \eqref{eq:spectral_asymptote} gives $\epscrit \simeq \varepsilon_0\sqrt{\kappa B/\sigma} = 2\sqrt2\sqrt{\kappa B\,\zeta/\wnat}$, and by \eqref{eq:wnat}
\[
\frac{\zeta}{\wnat} = \frac{\etath a}{2\,\etath\etalam c^2} = \frac{a}{2\etalam c^2},
\]
in which the primal step size cancels, whence $\epscrit \simeq (2/c)\sqrt{\kappa Ba/\etalam}$. The mechanism behind the cancellation is that $\etath$ enters the damping $\sigma \simeq \etath a/2$ and the tongue width in the same proportion, so reducing it lowers the threshold and narrows the tongue by the same factor, and at fixed physical bandwidth the two effects offset. \Halmos
\end{proof}

The averaging argument fixes both ends of the interpolation \eqref{eq:spectral}, $\epscrit(0) = \varepsilon_0$ exactly by Proposition~\ref{prop:spectral} and $\epscrit(B) \simeq \varepsilon_0\sqrt{B/(\pi\sigma)}$ for $B \gg \sigma$ by the proof above, but it does not fix the crossover at intermediate $B$: the second-order averaging theorem is an asymptotic statement in the scale separation between $\sigma^{-1}$ and the correlation time of $m$, valid in the two limits and silent in between, where Figure~\ref{fig:tongue} shows the tongue itself changing shape with depth. The one-parameter form $\epscrit(B) = \varepsilon_0\sqrt{1 + \kappa B/\sigma}$ of \eqref{eq:spectral} is the simplest interpolation that is exact at both derived ends, standing independent of the averaging argument beyond those two points, and it is on that residual construction, separate from the value of $\kappa$ itself, that the measurement of Appendix~\ref{sec:exp_broadband} bears: at the four bandwidths of Table~\ref{tab:ec_sqrtB}, the derived law $\kappa = 1/\pi \approx 0.3183$ predicts thresholds of $0.1372$, $0.1630$, $0.2401$ and $0.4263$ at $B/\sigma = 0.5$, $2$, $8$ and $32$ (using the measured single-line threshold $\varepsilon_0 = 0.12744$ and the interpolation \eqref{eq:spectral}), against the measured values of Table~\ref{tab:ec_sqrtB}, and the constant deduced independently from each band, $\kappa = 0.309 \pm 0.029$, brackets the derived value within a third of its own standard deviation.

\subsection{Existence corollary}\label{sec:ec_proof_existence}

\begin{proof}{Proof of Corollary~\ref{cor:existence}}
The depths in \eqref{eq:modulation} satisfy $|\varepsilon| \le 1$, since $|\varepsilon| > 1$ reverses the sign of a sensitivity and is not a perturbation of the same problem; hence $|h| \le 2$, with equality only if both sensitivities vanish at some point of the cycle. The first-order threshold $4\zeta = (2a/c)\rratio^{-1/2}$ crosses $2$ at $\rratio = (a/c)^2$, which is \eqref{eq:existence}. Corollary~\ref{cor:existence} separates the two halves of this statement by status. \Halmos
\end{proof}

The upper half of that status split, between $(a/c)^2/4$ and $(a/c)^2$, rests on a threshold evaluated at the centre of the tongue of the \emph{unmodulated} system; at large mismatch the tongue tilts and its maximum moves, so that threshold alone overestimates the stability region and pushes the true existence boundary below $(a/c)^2$, by an amount quantified empirically in Appendix~\ref{sec:ec_controls} (Table~\ref{tab:borkar_refine}). Growth rates just past that true boundary are minute in absolute terms.

\subsection{A first-order critic lag}\label{sec:ec_proof_lowpass}

\begin{example}[A first-order critic lag]
Take literally the motivating asymmetry of Section~\ref{sec:setup_channels}: the dual reads the true sensitivity without delay, $c_{\lambda,t} = c(1 + \varepsilon\sin\omega t)$, so $\epslam = \varepsilon$, while the primal's sensitivity estimate $c_{\theta,t}$ is produced by a critic that has to learn how the constraint responds to the policy, and so tracks the true sensitivity through first-order training dynamics, $\dot c_\theta = (c_{\lambda,t} - c_\theta)/\tau$ for some time constant $\tau$; this is a lag on the sensitivity itself, distinct from the constraint level $b_t$, which a critic observes directly and which plays no role in the parametric channel of Definition~\ref{def:channels}. At the tongue frequency $\omega = 2\wex$, the lag contributes the standard complex gain $H(i\omega) = 1/(1 + i\omega\tau)$, so the primal path sees a sensitivity modulation of depth $|\epsth| = \varepsilon/\sqrt{1 + (\omega\tau)^2}$ lagging the dual's by $\Delta\varphi = \arctan(\omega\tau)$, that is $\epsth e^{i\Delta\varphi} = \varepsilon H(i\omega)$. By the complex form of the threshold (Section~\ref{sec:parametric}), the mismatch entering it is
\begin{equation}
|\epslam - \epsth e^{i\Delta\varphi}| = \varepsilon\,|1 - H(i\omega)| = \varepsilon\,\frac{\omega\tau}{\sqrt{1 + (\omega\tau)^2}},
\end{equation}
which vanishes as the critic tracks faster than the tongue period, $\omega\tau \to 0$, recovering coherent cancellation without either channel being told to match the other, and saturates at the raw depth $\varepsilon$ as the critic falls arbitrarily far behind, $\omega\tau \to \infty$, the case already noted as yielding no protection. Instability by \eqref{eq:threshold} requires $\varepsilon\,\omega\tau/\sqrt{1+(\omega\tau)^2} > 4\zeta$, solved by $\omega\tau > 4\zeta/\sqrt{\varepsilon^2 - 16\zeta^2}$, which has a solution only if $\varepsilon > 4\zeta$: at the admissible depth $\varepsilon = 1$, a critic run slower than that bound is destabilizing, and none is once $\zeta \ge 1/4$. That existence condition is tighter than \eqref{eq:existence} by a factor $2$ in $\zeta$, $\zeta < 1/4$ here against $\zeta < 1/2$ there, because filtering one channel to zero reaches a mismatch of $\varepsilon$, half of the $2\varepsilon$ that two independently modulated channels of opposite sign can reach; since $\rratio$ scales as $\zeta^{-2}$, the same tightening is a factor $4$ in $\rratio$.
\end{example}

\section{Additional experiments}\label{sec:ec_more_exp}

\subsection{The additive channel}\label{sec:exp_additive}

Table~\ref{tab:additive} compares the coupled response to the frozen-policy response over 6 decades of forcing frequency, at $\etath = 0.05$ and $\etalam = 0.01$, for which Proposition~\ref{prop:floor} predicts a tracking plateau at $2Aa/c^2 = 0.2$ and a corner at $\omega_c = \etalam c^2/a = 10^{-2}$: the plateau value is $0.2$ at $\omega \le 10^{-4}$, the ratio to the frozen-policy amplitude, which diverges as $2\etalam A/\omega$ since a frozen policy cannot track the moving optimum $\lambda^\star(t)$, reaching 4 orders of magnitude at $\omega = 10^{-6}$.

Proposition~\ref{prop:worstcase} is tested in Table~\ref{tab:ec_worstcase} by maximizing the peak-to-peak amplitude over the forcing frequency across 3 decades of timescale ratio: the leading-order reduction $\max\{1, \rratio c^2/a^2\}$ is recovered within a factor of $1.00$ to $1.47$, consistent with the retained $\mathcal O(\zeta^2)$ correction of the resonant-peak asymptotic.

\subsection{The parametric channel under broadband forcing}\label{sec:exp_broadband}

Table~\ref{tab:ec_coherent_band} measures a coherent modulation spread over a band. Part (i) of Proposition~\ref{thm:coherent} already settles the sign at any frequency and waveform; the band changes only the size of the degradation, and does so as predicted, moving back towards $-\sigma$ as the band widens.

\begin{figure}
\FIGURE
{\includegraphics{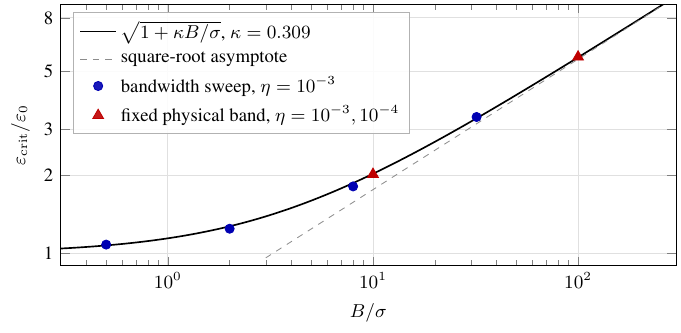}}
{Bandwidth law: mismatch threshold normalized by its own single-line threshold\label{fig:bandwidth}}
{Circles are the bandwidth sweep at $\etath = 10^{-3}$; triangles are the two fixed-physical-bandwidth points at $\etath = 10^{-3}$ and $10^{-4}$; each point is divided by its own single-line threshold (raw numbers and predictions in Tables~\ref{tab:ec_sqrtB} and~\ref{tab:ec_etath}). The solid curve is \eqref{eq:spectral} with the measured $\kappa = 0.309$ as its only fitted quantity; the dashed line is its asymptote $\sqrt{\kappa B/\sigma}$. Six measurements spanning a factor $200$ in $B/\sigma$ and $10$ in $\etath$ fall on one curve.}
\end{figure}

Two experiments measure the law \eqref{eq:spectral} itself, plotted together in Figure~\ref{fig:bandwidth}. The first sweeps the bandwidth over a factor $64$: the threshold grows with the band as Proposition~\ref{prop:spectral} predicts, and the constant deduced independently from each band gives $\kappa = 0.309 \pm 0.029$, confirming $\kappa = 1/\pi \approx 0.318$. The second tests that the primal step size drops out once the bandwidth is fixed in physical units: two operating points a decade apart in $\etath$ agree within the estimator's draw-to-draw scatter (full numbers and predictions in Tables~\ref{tab:ec_sqrtB} and~\ref{tab:ec_etath}).

The narrowband collapse as $\sqrt{\etath/\etalam}$ therefore holds only up to $B \approx 3.2\,\sigma$; beyond it the threshold is $(2/c)\sqrt{\kappa Ba/\etalam}$, still destabilizing but at a rate raised by $\sqrt{B/\sigma}$.

\subsection{Extinguishing the instability, and what the projection does}\label{sec:exp_prescription}

The prescription of Section~\ref{sec:algo} is demonstrated by Table~\ref{tab:ec_recursion}: at equal modulation depth, a coherent signal contracts over the horizon while a dual-only mismatch grows, 6 orders of magnitude apart. Setting $\ghat = g$ is what moves a run from the second regime to the first, and changes nothing else in the algorithm.

With the projection $\proj{\cdot}$ active (Table~\ref{tab:ec_recursion}), the instability saturates into a bounded limit cycle rather than an unbounded divergence. An integrated regret on $J$ separates the two regimes more informatively than the clipped-step fraction: it rises monotonically with mismatch depth at the reference point but inverts a decade below it, consistent with the tolerance collapsing as $\sqrt{\etath/\etalam}$ without vanishing. The provable quantity throughout is the sign of the exponent, independent of amplitude.

\paragraph{The PID agent of Section~\ref{sec:exp_deeprl}.} At the tongue centre the forced response of the multiplier is $A_\omega = (2/3)\,\varepsilon\bar\theta\sqrt{\etalam/k_\lambda}$. Simulated without the projection, the identified Lagrangian loop follows the formula, $1.03$, $0.63$ and $0.50$ at $\omega = 0.022$, $0.030$ and $0.036$ against $1.04$, $0.62$ and $0.49$, and the identified PID loop locks in every seed at $\omega = 0.026$; with the additive input and the projection both present it locks in $4$ seeds out of $10$, and removing the additive input alone restores $7$. In the confirmation lot, run at $\omega \in \{0.022, 0.026\}$ (Table~\ref{tab:deeprl_cost}), the projection is active at $\omega = 0.026$ for $12\%$ and $19\%$ of the steps at $\varepsilon = 0.3$ and $0.35$, against $5\%$ and $10\%$ in the simulated loop, which locks $9$ seeds out of $20$ where the agent locks $2$. The simulated loop with projection also reproduces the widening of the Lagrangian tongue at $\omega = 0.018$ and $0.022$, with $4$ and $10$ seeds locked against $5$ and $7$ observed, and misses it at $\omega = 0.030$, with none against $6$.

\subsection{Limitations of the experiments}\label{sec:ec_limits}

The active-projection regime is treated numerically only, at two operating points of the reduced loop and in the agent of Section~\ref{sec:exp_deeprl}, where it decides the outcome of the PID agent. The spectral shape is rectangular throughout Appendix~\ref{sec:exp_broadband}, whereas a more realistic $1/f$ traffic spectrum should give a more favourable threshold. The interpolation \eqref{eq:spectral}, anchored at both ends by the derived value $\kappa = 1/\pi$, still fixes its crossover shape at intermediate $B$ by construction. The exchange between a resonance and the constraint offset that $\ghat = g$ leaves behind (Section~\ref{sec:algo}) is a modeling question, external to the loop itself.

The cells at $\varepsilon = 0.8$ of Section~\ref{sec:exp_deeprl} run at $b = 0.15$, a deviation recorded before any of their data. The lower offset lowers $k_\lambda$ and with it $\wex$, so the gains were identified again at $b = 0.15$ and the Floquet predictions revised before the runs were read: the one frequency the revised loop finds unstable at $\rratio = 10$, $\omega = 0.018$, sits at $1.04$ times its $2\wex$. At $\rratio = 1$ the identified $k_a$ falls $41\%$ below its scaled value, outside the $\pm 30\%$ validity rule fixed with the predictions, so the stability predicted there, which the $50$ unlocked runs confirm, rests on gains outside that rule.

\section{Supporting tables}\label{sec:ec_tables}

\begin{table}
\TABLE
{Mismatch threshold against bandwidth, and the constant of \eqref{eq:spectral}\label{tab:ec_sqrtB}}
{\begin{tabular}{@{}lccccc@{}}
\hline\up
$B/\sigma$ & 0 & 0.5 & 2 & 8 & 32 \\ \hline\up
threshold (RMS) & 0.1264 & 0.1382 & 0.1590 & 0.2310 & 0.4262 \\
$/$ $\varepsilon_0$ & 0.99 & 1.08 & 1.25 & 1.81 & 3.34 \\
$\kappa$ deduced & n/a & 0.352 & 0.279 & 0.286 & 0.318 \down\\ \hline
\end{tabular}}
{Pure mismatch $\epsth = 0$, $\epslam$ swept, band centred on $2\wex$, operating point of Table~\ref{tab:floquet}, predicted single-line threshold $\varepsilon_0 = 0.12744$ in root-mean-square convention. These are the circles of Figure~\ref{fig:bandwidth}.}
\end{table}

\begin{table}
\TABLE
{Reference operating point of the parametric experiments\label{tab:opoint}}
{\begin{tabular}{@{}ll@{\qquad}ll@{}}
\hline\up
$\etath$ & $10^{-3}$ & $\etalam$ & $10^{-1}$ \\
$\rratio$ & $100$ & $\wnat$ & $0.01$ \\
$\wex$ (discrete) & $0.00999216$ & $\sigma$ (per step) & $4.502\times10^{-4}$ \\
$\zeta = \sigma/\wnat$ & $0.045020$ & $2\wex$ & $0.0199843$ \\
threshold (peak) & $0.18022$ & threshold (r.m.s.) & $0.12744$ \down\\ \hline
\end{tabular}}
{The principal tongue sits at twice the natural frequency, $2\wex$, hence at half the natural period ($\approx 630$ steps). The naive closed form of \eqref{eq:threshold_closed}, $a\rratio^{-1/2}/2c = 0.05$, drops the $\etalam c^2/a$ correction and is not the threshold: the peak threshold above, computed from the exact $\sigma$ of \eqref{eq:sigma_omega}, is $10\%$ below $4\times0.05$.}
\end{table}

\begin{table}
\TABLE
{Additive channel: worst case over forcing frequency against the timescale ratio\label{tab:ec_worstcase}}
{\begin{tabular}{@{}lcccccc@{}}
\hline\up
$\etath$ & $\etalam$ & $\rratio$ & $\zeta$ & $\sup_\omega \Delta\lambda$ & $/\,(2Aa/c^2)\max\{1,\rratio\}$ & $/\,(A\etalam/\sigma)$ \\ \hline\up
$10^{-3}$ & $10^{-3}$ & 1 & 0.500 & 0.294 & 1.47 & 1.468 \\
$10^{-3}$ & $10^{-2}$ & 10 & 0.157 & 2.118 & 1.06 & 1.049 \\
$10^{-3}$ & $10^{-1}$ & 100 & 0.045 & 22.322 & 1.12 & 1.005 \\
$10^{-2}$ & $10^{-1}$ & 10 & 0.143 & 2.320 & 1.16 & 1.049 \\
$10^{-1}$ & $10^{-2}$ & 0.1 & 1.648 & 0.200 & 1.00 & n/a \down\\ \hline
\end{tabular}}
{Reference parameters $a = c = b = 1$, $A = 0.1$, so plateau $= 2Aa/c^2 = 0.2$ and $\rratio c^2/a^2 = \rratio$. The last two columns are the residuals of the two forms of Proposition~\ref{prop:worstcase}: the leading-order reduction, whose residual carries the factor $1 + \etalam c^2/a$ and therefore does not vanish with $\zeta$, and the first form $A\etalam/\sigma$, which does. The last row is overdamped, where the resonant peak of the first form does not exist and the supremum is the plateau, so that row tests the floor of Proposition~\ref{prop:floor}(i), distinct from (ii). Of the remaining rows only $\zeta = 0.045$ sits well inside the leading-order-in-$\zeta$ hypothesis of the proposition; the rows at $\zeta = 0.143$, $0.157$ and $0.500$ are reported to show how the residual grows as $\zeta$ approaches the boundary of that hypothesis.}
\end{table}

\begin{table}
\TABLE
{Parametric channel: collapse of the mismatch tolerance with the timescale ratio\label{tab:ec_rsweep}}
{\begin{tabular}{@{}lcccc@{}}
\hline\up
$\etath$ & $\rratio = \etalam/\etath$ & measured threshold & predicted $4\sigma/\wex$ & ratio \\ \hline\up
$10^{-5}$ & $10^{4}$ & 0.01801 & 0.01800 & 1.00 \\
$10^{-4}$ & $10^{3}$ & 0.05694 & 0.05693 & 1.00 \\
$10^{-3}$ & $10^{2}$ & 0.1801 & 0.1802 & 1.00 \\
$10^{-2}$ & $10$ & 0.5678 & 0.5764 & 0.99 \\
$10^{-1}$ & $1$ & out of range & 2.075 & n/a \down\\ \hline
\end{tabular}}
{Fixed $\etalam = 0.1$, $a = c = 1$. Predicted column is $4\sigma/\wex$, evaluated exactly from \eqref{eq:sigma_omega}. Threshold read at the tongue centre $\omega = 2\wex$; see Table~\ref{tab:borkar} for the effect of maximizing over $\omega$ near the boundary.}
\end{table}

\begin{table}
\TABLE
{Existence boundary refined, and the exponent just past it\label{tab:borkar_refine}}
{\begin{tabular}{@{}lccc@{}}
\hline\up
$\etath\etalam$ & bracket for $\rratio^\star$ & $\rratio$ & exponent past threshold (per step) \\ \hline\up
$10^{-2}$ & $(0.85, 0.90]$ & -- & -- \\
$10^{-4}$ & $(0.96, 0.98]$ & $0.98$ & $+3.4\times10^{-5}$ \\
$10^{-4}$ & $(0.96, 0.98]$ & $1$ & $+8.4\times10^{-5}$ \\
$10^{-4}$ & $(0.96, 0.98]$ & $10^{2}$ & $+4.6\times10^{-3}$ \down\\ \hline
\end{tabular}}
{The first row repeats the crossing sweep of Table~\ref{tab:borkar} at $\etath\etalam = 10^{-2}$, in place of $10^{-4}$, refining the bracket for $\rratio^\star$. The last three rows are the exponent at the largest admissible mismatch just past the crossing, at the ratios of Table~\ref{tab:borkar}, corresponding to doubling times of $20\,600$, $8\,300$ and $152$ steps.}
\end{table}

\begin{table}
\TABLE
{Coherent modulation under broadband forcing: top Lyapunov exponent per step\label{tab:ec_coherent_band}}
{\begin{tabular}{@{}lcc@{}}
\hline\up
$B/\sigma$ & $\varepsilon = 0.3$ & $\varepsilon = 0.7$ \\ \hline\up
0 (single line) & $-3.41\times10^{-4}$ & $-2.06\times10^{-4}$ \\
1 & $-3.91\times10^{-4}$ & $-2.68\times10^{-4}$ \\
4 & $-4.26\times10^{-4}$ & $-3.56\times10^{-4}$ \\
16 & $-4.43\times10^{-4}$ & $-4.06\times10^{-4}$ \down\\ \hline
\end{tabular}}
{Coherent modulation, $\epsth = \epslam = \varepsilon$ in root-mean-square convention, band centred on $2\wex$, operating point of Table~\ref{tab:floquet}, unmodulated exponent $-\sigma = -4.502\times10^{-4}$. Widening the band moves the exponent back towards $-\sigma$.}
\end{table}

\begin{table}
\TABLE
{Cancellation of the primal step size at fixed physical bandwidth\label{tab:ec_etath}}
{\begin{tabular}{@{}lcccccc@{}}
\hline\up
$\etath$ & $\rratio$ & $\sigma$ & $B/\sigma$ & single line & band $B$ & predicted band $B$ \\ \hline\up
$10^{-3}$ & 100 & $4.502\times10^{-4}$ & 10 & 0.1277 & 0.2570 & 0.2576 \\
$10^{-4}$ & 1000 & $4.500\times10^{-5}$ & 100 & 0.04222 & 0.2393 & 0.2273 \down\\ \hline
\end{tabular}}
{Fixed $\etalam = 0.1$ and fixed physical bandwidth $B = 4.5\times10^{-3}$ radians per step, thresholds in root-mean-square convention, prediction from \eqref{eq:spectral}. Ratio between the 2 rows: $3.02$ at a single line, $1.07$ at fixed physical bandwidth. These are the triangles of Figure~\ref{fig:bandwidth}.}
\end{table}

\begin{table}
\TABLE
{Full recursion over 200 periods (62\,800 steps), without and with the projection\label{tab:ec_recursion}}
{\resizebox{\textwidth}{!}{\begin{tabular}{@{}lcrrrrrr@{}}
\hline\up
$\epsth$ & $\epslam$ & Floquet exponent & $\exp(\mu T)$ & $\max\lambda$ (unprojected) & $\max\lambda$ (projected) & final $\Delta\lambda$ & \% clipped \\ \hline\up
0 & 0 & $-4.502\times10^{-4}$ & $5.26\times10^{-13}$ & 1 & 1 & 0 & 0.0 \\
0.3 & 0.3 & $-3.748\times10^{-4}$ & $5.98\times10^{-11}$ & 4.663 & 4.168 & 2.780 & 26.7 \\
0.9 & 0.9 & $-2.271\times10^{-4}$ & $6.38\times10^{-7}$ & 11.21 & -- & -- & -- \\
0 & 0.3 & $+2.988\times10^{-4}$ & $1.41\times10^{8}$ & $4.45\times10^{6}$ & 3.933 & 2.685 & 28.0 \\
0.3 & 0 & $+2.988\times10^{-4}$ & $1.41\times10^{8}$ & $4.66\times10^{5}$ & -- & -- & -- \\
0.2 & 0.5 & $+2.961\times10^{-4}$ & $1.19\times10^{8}$ & $1.17\times10^{6}$ & 5.669 & 3.649 & 36.6 \down\\ \hline
\end{tabular}}}
{Operating point of Table~\ref{tab:floquet}, forcing at $\omega = 2\wex$. The first 3 rows are coherent, the last 3 mismatched, and this is the comparison read in Appendix~\ref{sec:exp_prescription}. At depth $0.3$, 6 orders of magnitude separate the multiplier excursion under a coherent modulation from the excursion under a mismatch of the same depth. The two amplitude columns compare the unprojected recursion against the projection $\proj{\cdot}$ active on 4 of the 6 configurations (a dash marks a configuration not run under projection); the projected exponents are those of the linearized system, reproduced from the unprojected columns, while the projected amplitude, final $\Delta\lambda$, and clipped-step-fraction columns are measured on the projected recursion, where they are set by the non-linearity, independent of the exponent. The clip prevents the negative half of the diverging cycle, so the instability saturates into a limit cycle and the linear model is conservative on amplitude. Appendix~\ref{sec:exp_prescription} reads the clipped-step-fraction column.}
\end{table}

Integrated regret $\frac1T\sum_t \frac a2(\theta_t-\theta^\star)^2$ on the projected recursion, measured at the reference operating point of Table~\ref{tab:ec_recursion} and at a decade below it in $\rratio$ with \texttt{projected\_regime\_metrics.py}, gives a mismatch-to-coherent ratio of $1.14$ at depth $0.3$ and $2.15$ at depth $0.9$ for $\rratio = 10^2$, against $0.69$ to $0.86$ over depths $0.1$ to $0.9$ for $\rratio = 10$. Appendix~\ref{sec:exp_prescription} reads this comparison: the separation widens with depth at the reference point and inverts a decade below it, which is why the paper reads the sign of the exponent, independent of any amplitude, as the provable quantity.

\begin{table}
\TABLE
{Lagrangian SAC at $\rratio = 100$: locking, forced response, projection, return and constraint value against the forcing frequency\label{tab:deeprl_cost}}
{\begin{tabular}{@{}llccccc@{}}
\hline\up
 & $\omega$ & 0.018 & 0.022 & 0.026 & 0.030 & 0.036 \\
 & $\omega/2\wex$ & 0.71 & 0.87 & 1.03 & 1.19 & 1.43 \\ \hline\up
locked seeds & Lagrangian & 5/10 & 7/10 & 10/10 & 6/10 & 1/10 \\
 & PID & 0/10 & 1/10 & 0/10 & 1/10 & 0/10 \\
 & coherent & 0/5 & 0/5 & 0/5 & 0/5 & 0/5 \\
 & PID, $\varepsilon = 0.3$ & -- & 1/10 & 2/10 & -- & -- \\
 & PID, $\varepsilon = 0.35$ & -- & 1/10 & 0/10 & -- & -- \\ \hline\up
$A_\omega$ & predicted, unprojected & 1.67 & 1.04 & 0.77 & 0.62 & 0.49 \\
 & Lagrangian & 0.814 & 0.701 & 0.625 & 0.565 & 0.498 \\
 & PID & 0.821 & 0.737 & 0.686 & 0.621 & 0.529 \\
 & coherent & 0.866 & 0.810 & 0.732 & 0.606 & 0.489 \\
 & PID, $\varepsilon = 0.3$ & -- & 0.696 & 0.611 & -- & -- \\
 & PID, $\varepsilon = 0.35$ & -- & 0.718 & 0.657 & -- & -- \\ \hline\up
fraction $\lambda = 0$ & Lagrangian & 0.34 & 0.32 & 0.25 & 0.14 & 0.04 \\
 & PID & 0.34 & 0.29 & 0.23 & 0.18 & 0.05 \\
 & coherent & 0.30 & 0.20 & 0.11 & 0.05 & 0.03 \\
 & PID, $\varepsilon = 0.3$ & -- & 0.22 & 0.12 & -- & -- \\
 & PID, $\varepsilon = 0.35$ & -- & 0.26 & 0.19 & -- & -- \\ \hline\up
return & Lagrangian & 0.2107 & 0.2185 & 0.2270 & 0.2382 & 0.2459 \\
 & PID & 0.2168 & 0.2248 & 0.2321 & 0.2402 & 0.2465 \\
 & coherent & 0.2153 & 0.2294 & 0.2381 & 0.2432 & 0.2447 \\
 & PID, $\varepsilon = 0.3$ & -- & 0.2357 & 0.2428 & -- & -- \\
 & PID, $\varepsilon = 0.35$ & -- & 0.2302 & 0.2376 & -- & -- \\ \hline\up
$10^3\,\bar g$ & Lagrangian & $-39.9$ & $-31.0$ & $-22.0$ & $-10.1$ & $-2.5$ \\
 & PID & $-36.7$ & $-26.9$ & $-18.0$ & $-8.1$ & $-1.5$ \\
 & coherent & $-33.3$ & $-17.1$ & $-7.2$ & $-2.3$ & $-1.6$ \\
 & PID, $\varepsilon = 0.3$ & -- & $-13.8$ & $-5.3$ & -- & -- \\
 & PID, $\varepsilon = 0.35$ & -- & $-20.4$ & $-11.4$ & -- & -- \down\\ \hline
\end{tabular}}
{Agent and operating point of Section~\ref{sec:exp_deeprl}, $\varepsilon = 0.4$ unless stated, $b = 0.3$, $2\wex = 0.0252$; the last two rows of each block are the confirmation lot, a dash marking a frequency not run. Second half of each $20\,000$-step run: $A_\omega$ is the amplitude of $\lambda$ at the forcing frequency and the fraction of steps at $\lambda = 0$ the activity of the projection, both medians over seeds; the return of the deterministic action and the windowed constraint value $\bar g$ are means over seeds, $\bar g < 0$ meaning a slack constraint. The predicted row is $\etalam\varepsilon\bar\theta\,\omega/|\omega^2 - \wex^2|$ with $\etalam = 0.132$ and $\bar\theta = 0.291$, to compare with $\lambda^\star = 0.70$. Unmodulated runs return $0.2480$ (Lagrangian) and $0.2485$ (PID).}
\end{table}

\section{Instrument controls}\label{sec:ec_controls}

This section collects the validation runs quoted in Section~\ref{sec:experiments}. They are reported in this appendix because none of them measures the phenomenon: each one checks that an instrument returns a value that is known in advance, so that the measurements of Section~\ref{sec:experiments} can be read as genuine measurements, free of artifacts.

\paragraph{Additive channel.} Three controls support Table~\ref{tab:additive}. Simulating the full recursion including the projection reproduces the analytic values to better than $1.2\%$ at $\omega = 0.3$, a frequency picked for this control independently of the sweep tabulated in Table~\ref{tab:additive}, and to better than $10^{-4}$ relative for $\omega \le 0.03$, so the plateau is not an artifact of linearization. The plateau value measured at $\omega = 10^{-8}$ is $0.200000$ across a four-decade range of each step size, $\etalam \in [10^{-4}, 10^{-1}]$ and $\etath \in [10^{-3}, 0.5]$, confirming the independence from both step sizes asserted by Proposition~\ref{prop:floor}. And the frozen-policy column agrees with the analytic law $\etalam A/|\sin(\omega/2)|$ to $10^{-9}$ relative, which validates the instrument against the independently derived bound of \citet{baelde2026companion}.

\paragraph{Periodic parametric channel.} The Floquet measurements of Section~\ref{sec:exp_narrowband} are bracketed by 4 controls. The first isolates the coupling: at frozen policy, $\etath = 0$, the Floquet exponent is exactly zero for every modulation depth including $\varepsilon = 0.9$, so the parametric failure mode does not exist outside the coupled system, which is the mirror image of the additive result where the frozen policy diverged and the coupling stabilized. The second validates the instrument: at $\varepsilon = 0$ the exponent returns $\log\rho(M) = -4.502\times10^{-4}$, the closed-form value of \eqref{eq:sigma_omega}. The third localizes the phenomenon: at fixed mismatch $\epslam = 0.3$, $\epsth = 0$, the exponent equals that of the unmodulated system at $\omega/2\wex \in \{0.4, 0.7, 1.5, 2.5\}$ and is positive only at $\omega/2\wex = 1$, confirming a localized resonance, without a generic degradation of stability. The fourth modulates the curvature $a$ in place of the sensitivity, at depth $\varepsilon_a$, and finds stability even at depth $0.9$, exponent $-2.25\times10^{-4}$, matching the first-order prediction $-\sigma(1-\varepsilon_a/2)$ from the pure damping modulation this induces; a curvature modulation destabilizes only at depths above the admissible range of the sensitivity modulation, $\varepsilon_a = 2(1 - \etalam c^2/a)$.

\paragraph{The existence boundary's two normalizations.} Table~\ref{tab:borkar} reads its admissible bound against two forms of the first-order threshold that agree only to relative order $\zeta^2$. For the same bound the two threshold columns, $4\sigma/\wex$ and $4\sigma/\wnat$, place the predicted crossing at $\rratio = 1.2476$ and $\rratio = 1.0098$ respectively, and the continuous form $(2a/c)\sqrt{\etath/\etalam}$ from which \eqref{eq:existence} is read places it at $\rratio = 1.0203$, a spread of $24\%$ that exceeds the distance from any one of them to the measured bracket $(0.96, 0.98]$. Read as a design boundary, the first-order value is optimistic in the unsafe direction at both products, by $3\%$ under $4\sigma/\wnat$ and by $27\%$ under $4\sigma/\wex$ at $\etath\etalam = 10^{-4}$, and the drift of that gap with the step sizes changes sign between the two normalizations.

\paragraph{Broadband parametric channel.} The Lyapunov instrument needs its own controls before it can be trusted off the periodic case. With no modulation it returns $-4.5357\times10^{-4}$ against the closed-form $-\sigma = -4.5020\times10^{-4}$. The gap of $3.4\times10^{-6}$ is a finite-window bias: the eigenvalues are complex, the norm of the iterate oscillates, and averaging over a non-integer number of periods leaves an $\mathcal O(1/T)$ residue. It biases the exponent, leaving the threshold's position, set by the sign change and dominated by a second effect, unaffected. With a single line, a sinusoid of random phase at pure mismatch $\epsth = 0$, the measured root-mean-square threshold is $0.1274$ against the predicted $\varepsilon_0 = 0.1274$. That agreement is quoted on one draw of the signal, and the estimator has a scatter of its own: bisected to $10^{-5}$ on 5 independent draws at this operating point, the threshold reads $0.12737$, $0.12815$, $0.12712$, $0.12595$ and $0.12776$, a mean of $0.12727$ with a standard deviation of $0.00084$, $0.66\%$, and a range of $1.7\%$, the prediction falling inside. The same check repeated at the four nonzero bandwidths of Table~\ref{tab:ec_sqrtB}, 50 bisection halvings against the 9-point interpolation on one and the same draw at each bandwidth, gives a relative gap rising from $0.02\%$ at $B/\sigma = 0.5$ to $1.12\%$ at $B/\sigma = 32$, and a bisected $\kappa = 0.326 \pm 0.040$ against the interpolated $0.324 \pm 0.040$ from that same nine-point grid, so the interpolation grid contributes negligibly to the scatter. So the Lyapunov machinery reproduces the Floquet threshold to within the scatter of its own estimator, and the several single-line values quoted across this paper are repeated draws of a single underlying measurement standing for one threshold. And with a band of width $2\sigma$ at strong mismatch, only the band centred on the tongue destabilizes: for a centre frequency at $0.4$, $0.7$, $1.0$, $1.6$ and $3.0$ times $2\wex$, the exponent is $-2.94$, $-3.93$, $+10.7$, $-4.53$ and $-4.48$ in units of $10^{-4}$.

\section{Numerical protocol}\label{sec:protocol}

For the reduced recursion all arithmetic is double precision and no additive gradient noise is injected, in keeping with the model of Section~\ref{sec:setup}, additive gradient noise being treated separately in Remark~\ref{rem:multiK}. The bare numerical parameters of every sweep and estimator below are collected in Table~\ref{tab:ec_protocol_params}; the text states only what each choice is for and how it was validated.

\begin{table}
\TABLE
{Numerical parameters of the protocol\label{tab:ec_protocol_params}}
{\begin{tabular}{@{}lp{0.5\linewidth}@{}}
\hline\up
Additive control (Table~\ref{tab:additive}) & $40$ forcing periods \\
Additive worst-case sweep (Table~\ref{tab:ec_worstcase}) & $4000$ points log-spaced, $\omega \in [10^{-7}, \pi]$ \\
Floquet tongue-centre maxima & $121$ points on $\omega/2\wex \in [0.4, 1.6]$, $35$ on $[1.6, 5.0]$ \\
Figure~\ref{fig:tongue} grid & $589$ integer periods, $196$--$784$ steps \\
Tongue-centre mismatch bisection & $50$ halvings \\
Existence-boundary sweep (Table~\ref{tab:borkar}) & $60$ periods log-spaced, $8$--$3000$ steps, refined by $121$ points on $[0.55,1.45]\,\pi/\wex$ \\
Existence-boundary bisection & $18$ halvings, mismatch split $\epsth = +d/2$, $\epslam = -d/2$, $d \le 1.98$ \\
Broadband Lyapunov estimator & $T = 200\,000$ steps, $16$ realizations, $40\,000$-step burn-in \\
Broadband bandwidth sweep & $B/\sigma \in \{0, 0.5, 2, 8, 32\}$, 9-point mismatch grid $\{0.05,\ldots,1.0\}$ \\
Primal-step-size comparison & $T = 400\,000$ steps, $8$ realizations, burn-in $T/5$, 9-point grid on $[0.4\varepsilon_0, 10\varepsilon_0]$ \\
Random seed & $20\,260\,821$ (NumPy PCG64, one draw per script run) \down\\ \hline
\end{tabular}}
{$\varepsilon_0 = 2\sqrt2\zeta$ is the pure-line threshold of the row it is quoted against, so the primal-step-size grid follows the prediction, in place of being held fixed while $\zeta$ changes.}
\end{table}

\begin{table}
\TABLE
{Additive channel: peak-to-peak multiplier excursion tracking the moving optimum, coupled against frozen policy\label{tab:additive}}
{\begin{tabular}{@{}lccc@{}}
\hline\up
$\omega$ & coupled & frozen policy & ratio \\ \hline\up
$10^{-6}$ & 0.200000 & 2000.000 & 10\,000 \\
$10^{-4}$ & 0.199994 & 20.000 & 100 \\
$10^{-3}$ & 0.199450 & 2.000 & 10 \\
$10^{-2}$ & 0.159698 & 0.200 & 1.3 \\
$10^{-1}$ & 0.020894 & 0.020008 & 1.0 \\
$1$ & 0.002087 & 0.002086 & 1.0 \down\\ \hline
\end{tabular}}
{Reference parameters $a = c = b = 1$, $A = 0.1$, $\etath = 0.05$, $\etalam = 0.01$. Predicted plateau $2Aa/c^2 = 0.2$, predicted corner $\omega_c = 10^{-2}$.}
\end{table}

\paragraph{Additive channel.} Frequency responses are the closed-form peak-to-peak amplitude read from \eqref{eq:transfer}. Each entry of Table~\ref{tab:additive} has its control in direct simulation of the full projected recursion, with the peak-to-peak measured on the last period alone so that the transient is excluded; the frozen-policy column is the same simulation with the primal update disabled. The supremum over the forcing frequency reported in Table~\ref{tab:ec_worstcase} sweeps up to the Nyquist frequency of the recursion. The plateau and corner values quoted in the text are the analytic limits of \eqref{eq:transfer}, verified against the grid to a relative $10^{-3}$.

\paragraph{Periodic parametric channel.} The forcing period is always an integer number of steps, $N = \max\{8, \operatorname{round}(2\pi/\omega)\}$, so the monodromy $\Phi = \prod_{t<N}M_t$ closes exactly and the Floquet exponent $\log\rho(\Phi)/N$ carries no truncation error. The implementation is controlled at $\epsth = \epslam = 0$, where $\Phi$ must equal $M^N$ to machine precision. Exponents at the tongue centre are read at $N$ corresponding to $\omega = 2\wex$; the refinement below $1$ in the tongue-centre sweep is what captures the tongue tilt discussed in Corollary~\ref{cor:existence}. Figure~\ref{fig:tongue} sweeps integer periods, in place of frequencies, which removes the rounding of $2\pi/\omega$ altogether, at a frequency resolution that varies across the range, $0.3\%$ near the centre of the principal tongue. The existence-boundary sweep of Table~\ref{tab:borkar} cannot use a grid relative to $\wex$, since the overdamped rows have no oscillatory mode at all, so it sweeps an absolute period grid instead; without the refinement near resonance the narrow tongues are missed and the measured threshold reflects the grid spacing, independent of the dynamics. The mismatch bound there is the largest reachable with $|\varepsilon| \le 0.99$ on each channel, and a threshold reported as absent means that no instability was found at that bound for any frequency on the grid.

\paragraph{Broadband parametric channel.} The modulation is synthesized by placing independent complex Gaussian coefficients on the lines of the real Fourier grid that fall inside $[\omega_c \pm B/2]$, zeroing every other line, and inverse transforming; each realization is then normalized to unit variance, so $\varepsilon$ is a root-mean-square depth. A band carrying fewer than 3 lines is rejected as under-resolved, in place of being silently degenerated. The single-line case, which reproduces the periodic setting with a random phase, is retained as a control. Top Lyapunov exponents are estimated by iterating a renormalized vector, with the reported spread being the standard deviation across realizations; the estimator is controlled at zero modulation, where it must return $-\sigma$ to within $10^{-5}$. Thresholds are obtained by linear interpolation of the first sign change on the mismatch grid. Each realization is a finite Fourier synthesis, a multisinusoid with independent random phases on the retained lines: it is wide-sense stationary, in place of the strictly stationary, ergodic process invoked in the proof of Proposition~\ref{prop:spectral} (Appendix~\ref{sec:ec_proof_lyapunov}), and the exponent is estimated on that finite-length surrogate; the estimator control at zero modulation and the agreement with the Floquet threshold at $B = 0$ show no dependence on this distinction at the horizons used here.

\paragraph{Full recursion.} Table~\ref{tab:ec_recursion} simulates the unlinearized recursion of Section~\ref{sec:setup_dynamics}, affine term included, at the reference operating point for $200$ forcing periods, that is $62\,800$ steps. The initial condition is the unmodulated saddle $(\theta_0,\lambda_0) = (-b/c,\, ab/c^2)$, which is $(-1, 1)$ under the parameterization used here, so the unmodulated row reports $\max_t|\lambda_t| = 1$ and the departure from that value measures the excursion. The final peak-to-peak is measured on the last period alone, and the clipped fraction is the proportion of steps at which the unprojected update would have sent $\lambda$ negative, which is counted in both runs and acted upon only in the projected one. A divergence guard freezes a run at $|\lambda_t| > 10^{13}$; the largest excursion reported is 6 orders of magnitude below it, so the guard never fires in that table.

\paragraph{Code availability.} The tables above are reproduced by 7 self-contained scripts, depending only on NumPy and publicly available, with the agent code, at \url{https://github.com/mbaelde/dualdynamics-code}, each carrying the controls described above as executable assertions so that a run completing silently has also verified its own instruments. Four cover the additive channel and its frequency responses, the periodic parametric channel and its Floquet diagnostics, the broadband channel and its Lyapunov diagnostics, and the existence-boundary sweep across the timescale ordering. A fifth produces the data behind Figures~\ref{fig:tongue} and~\ref{fig:bandwidth}: it calls the Floquet instruments of the third one for Figure~\ref{fig:tongue}, which is therefore a fresh measurement whose value at $\omega = 2\wex$ coincides with Table~\ref{tab:floquet} by construction, and it forms the dimensionless ratios of Figure~\ref{fig:bandwidth} from the thresholds of the bandwidth sweep and of the primal-step-size comparison, so that figure re-presents tabulated measurements, without constituting an independently generated run. A sixth, \texttt{projected\_regime\_metrics.py}, produces the integrated regret figures of Appendix~\ref{sec:exp_prescription}. A seventh, \texttt{toy\_phase\_and\_multiK.py}, verifies numerically, on the tangent reduction, the phase-offset generalization of the threshold discussed after Lemma~\ref{lem:cancellation}.

\paragraph{Deep RL agent.} The agent of Section~\ref{sec:exp_deeprl} is outside the protocol above. The actor is a tanh-squashed Gaussian and each critic a regression of the immediate reward or cost, all with two hidden layers of $64$ ReLU units; the episodes have one step, so the critics need no target network. The critics are trained jointly by Adam at step size $10^{-3}$, the actor by SGD, one gradient step each per environment step on minibatches of $256$ from the replay buffer, after $1\,000$ uniformly random warm-up actions, with entropy temperature fixed at $0.01$ and $\lambda$ started at $0.7$. The actor and dual step sizes are $(3\times10^{-3}, 0.1)$ at the reference point and $(2.28\times10^{-2}, 0.0132)$, $(7.20\times10^{-3}, 0.0417)$ and $(2.28\times10^{-3}, 0.132)$ at $\rratio = 1$, $10$ and $100$, all with the product $3\times10^{-4}$. Each cell uses seeds $1$ to $10$, $1$ to $5$ for the coherent agent. The amplitudes $A_{1/2}$ and $A_\omega$ are lock-in estimates $2|\langle\lambda_t e^{-i\nu t}\rangle|$ over the second half of the run, at $\nu = \omega/2$ and $\nu = \omega$. The scripts that train the agent, identify its loop and produce Figure~\ref{fig:deeprl} and Table~\ref{tab:deeprl_cost} depend on PyTorch and are in the same repository as the seven above.

\end{document}